\documentclass[submit]{aiaa-pretty}
\usepackage{amssymb,array,enumerate,epsfig,pifont,graphics,subfigure,paralist,amsmath}
\usepackage{algpseudocode}
\usepackage{algorithm}
\usepackage{setspace}
\usepackage{graphicx}
\usepackage{transparent}
\usepackage{color}
\usepackage[titletoc]{appendix}
\usepackage{gensymb}
\usepackage{xparse}
\usepackage{indentfirst}
\usepackage{gensymb}

\usepackage{afterpage}

\usepackage{multirow}

\usepackage{csquotes}

\usepackage{amsthm}
\newtheorem{mydef}{Definition}

\usepackage[normalem]{ulem}

\usepackage{fullpage}

\author[Abhishek Verma \ and B\'{e}r\'{e}nice Mettler]{ %
Abhishek Verma 
and
B\'{e}r\'{e}nice Mettler
}

\title{Human Learning of Unknown Environments in Agile Guidance Tasks}

\abstract{Trained human pilots or operators still stand out through their efficient, robust, and versatile skills in guidance tasks such as driving agile vehicles in spatial environments or performing complex surgeries. 
This research studies how humans learn a task environment for agile behavior. The hypothesis is that sensory-motor primitives previously described as interaction patterns and proposed as units of behavior for organization and planning of behavior provide elements of memory structure needed to efficiently learn task environments. 
The paper presents a modeling and analysis framework using the interaction patterns to formulate learning as a graph learning process and apply the framework to
investigate and evaluate human learning and decision-making while operating in unknown environments. This approach emphasizes the effects of agent-environment dynamics (e.g., a vehicle controlled by a human operator), which is not emphasized in existing environment learning studies. 
The framework is applied to study human data collected from simulated first-person guidance experiments in an obstacle field. Subjects were asked to perform multiple trials and find minimum-time routes between pre-specified start and goal locations without priori knowledge of the environment.}

\DeclareMathOperator*{\Min}{Min}

\begin{document}
\maketitle

\def\dd{\partial}
\def\bI{{\bf I}}

\section{Introduction}
\label{sec:introduction}

Humans are capable of learning complex unknown environments in a variety of guidance tasks and use the knowledge to determine near-optimal (e.g., minimum-time) performance and remain versatile and adaptive to unexpected changes in the environment. This capability is not unique to spatial environment navigation but is also essential to other spatial tasks involving interactions with the environment such as pertaining to surgery. The general goal of this research is to understand how humans achieve efficient environment learning and path-planning capabilities despite their limited sensing, information processing, and memory capabilities. The understanding of relevant cues humans use for environment learning and path planning can be used in: 1) autonomous guidance algorithms to 
be computationally efficient and adaptive to changes in the task environment, 2) human-machine interfaces to focus on cues that help human operators in environment learning and planning. 

An agile guidance task in an unknown environment primarily involves three steps. The first step is environment sensing and assimilating the sensed environment information into global knowledge. The second step is path planning, i.e., trajectory optimization, using the known/learned knowledge and planning an immediate trajectory. The last step is tracking the planned trajectory. The three steps are repeated online. An autonomous guidance operation requires a mechanism for sensing and learning the environment and representing the learned environmental information in computationally efficient ways, in order to process online trajectory planning. Given the limited sensing and information processing capabilities of autonomous guidance systems, it is challenging to develop efficient learning and representation methods for real-world tasks in spatial environments.

Humans have constraints on their memory, sensing, and information processing capabilities. Human limitations are limited field of view and visual attention, limited information processing (e.g., working memory), and perceptual guidance at sensing, planning, and control levels, respectively. Despite these limitations, they can navigate complicated unknown environments, exhibit efficient behavior in agile guidance tasks, and given enough trials, learn near-optimal solutions (e.g., minimum-time route between two places). Strengthening the knowledge base about environment learning in humans could be a key to overcoming computational complexities in autonomous guidance systems that arise from having to process enormous amount of information available in real-world task environments. This research investigates principles that underlie robust human guidance, learning  process/mechanism, and memory structure in dynamic spatial behavior/navigation of unknown environments.

Humans and other animals have evolved a system of processes to navigate and interact with their environments. Gibson~\cite{Gibson1979} introduced the idea that spatial behaviors are mediated by affordances, available through the interaction with task and environmental elements. For example, Lee~\cite{Lee1976,Lee1998,LeeKalmus1998} showed how principles like time-to-closure of a gap, e.g., distance, angle, force, etc., and optical flow can be used to regulate motion, rather than requiring complex models and computations. Tau-control and optical flow show how humans and animals use limited cues from the environment, which help them overcome their perceptual and information processing constraints. Inspired by the concept of affordances and limited cues, this research investigates cues and affordances used by humans to navigate unknown environments.

The central concept this research is based on is ``invariants". Simon~\cite{Simon1990Invariants} quoted ``The fundamental goal of science is to find invariants". An invariant can break down a complex problem into smaller subproblems such that a similar solution can be used for a set of subproblems. For guidance tasks in spatial environments, previous studies~\cite{kong2011investigation,KongHMSIEEE2013,MettlerFrontiers2015} with human pilots operating remote-control miniature rotorcraft showed that pilots organize spatial behavior by using invariants in their sensory-motor behavior (guidance, control, and perceptual processes) in interaction with the spatial environment and task elements. The invariants in sensory-motor behavior are called interaction patterns (IPs) as these emerge from interactions between the agent and the task environment. IPs are transferable to similar task domains via symmetry transformations such as rigid-body transformation (translation, rotation, and reflection), which mitigates a guidance task complexity. Mettler et. al~\cite{MettlerFrontiers2015} proposed that IPs function as units of organization for planning spatial behavior in guidance tasks. Previously, IPs have been studied and used for modelling human guidance behavior in known environments. This paper investigates what functions IPs play in human learning during goal-directed guidance tasks in unknown environments. The paper builds on IPs to propose a framework that allows to formally investigate human environment learning in agile guidance tasks.    

\subsection{Motivation: Engineering vs. Spatial Cognition}

Spatial cognition studies in general focus on pedestrians or simple movements. As stated by the author in \cite{Mettlercogncrit2011}, ``simple forms of navigation, or way finding, have been the main focus of spatial cognition but without accounting for the effects of dynamics." In agile guidance tasks, such as a pilot operating a high-speed vehicle in a complex environment or surgeons under time pressure, the interactions between vehicle dynamics and task environment play a role in determining what elements of the environment are more relevant than others.  

The overall behavior of a human pilot in a spatial environment is laid out by the interactions between pilot control and cognitive characteristics and the environmental characteristics such as scale and layout. 
Warren~\cite{Warren2006} used the term ``behavioral dynamics" that represents the closed-loop agent-environment dynamics. The concept is originally inspired from the Gibson's idea of ecological perception~\cite{Gibson1979}. Gibson's ecological approach suggests that a human or animal learns (represents) an environment based on the task in hand and desired goals. Therefore, the study of spatial cognition (representation and learning) should be integrated with the study of pilot dynamic and perceptual behavior. 


Mettler~\cite{Mettlercogncrit2011} highlighted that traditional optimal control formulation of trajectory planning problems does not take advantage of the problem structures that play a fundamental role in humans' and animals' skills. The author proposed the idea that skilled human pilots possess a system to conceptualize spatial behavior that preserves the interrelation between movement dynamics and geometry and topology of the environment. 
In subsequent studies, Kong and Mettler~\cite{KongHMSIEEE2013} studied the guidance behavior in complex environments focusing on the agent-environment interactions. The study showed that skilled operators organize their behavior according to interaction patterns. These sensory-motor patterns represent units of behavior which satisfy the various system constraints and exploit the equivalences in the problem space.
Furthermore, the interaction patterns make it possible to abstract a task environment as a graph of subgoals. Such graph framework can be elaborated to build a cognitive map to model and investigate human learning and decision-making in complex task environments. This paper uses the subgoal graph to investigate human environment learning and spatial navigation in guidance tasks where human subjects navigate using a complex dynamic vehicle. 

\subsection{Experiments and Data}
\label{subsec:experimentsystem}

This section gives an overview of the experiment system and human data used for the investigation of human environment learning. 

\subsubsection{Experiment System}

The guidance experiments were conducted on the system introduced in \cite{FeitSMC2015} (see in Fig.~\ref{SimSystem}(a)). The system consists of a monitor to display a simulated task environment, a joystick to control flight behavior and navigate in the environment, and a gaze tracking device to record 3-D gaze location. The system provides a first-person view with a limited field of view (60$\degree$) to human subjects. The longitudinal and lateral control inputs ($u_{lon}$ and $u_{lat}$, respectively) correspond to forward speed ($v$) and turn-rate ($\omega$). There is a delay between speed command $u_{lon}$ and vehicle speed $v$. Turn-rate is inversely proportional to the speed. Vehicle dynamic model is given in Section~\ref{sec:formulation}. 


\subsubsection{Experiments}

\begin{figure}[htbp]
\centering
\includegraphics[scale=0.4]{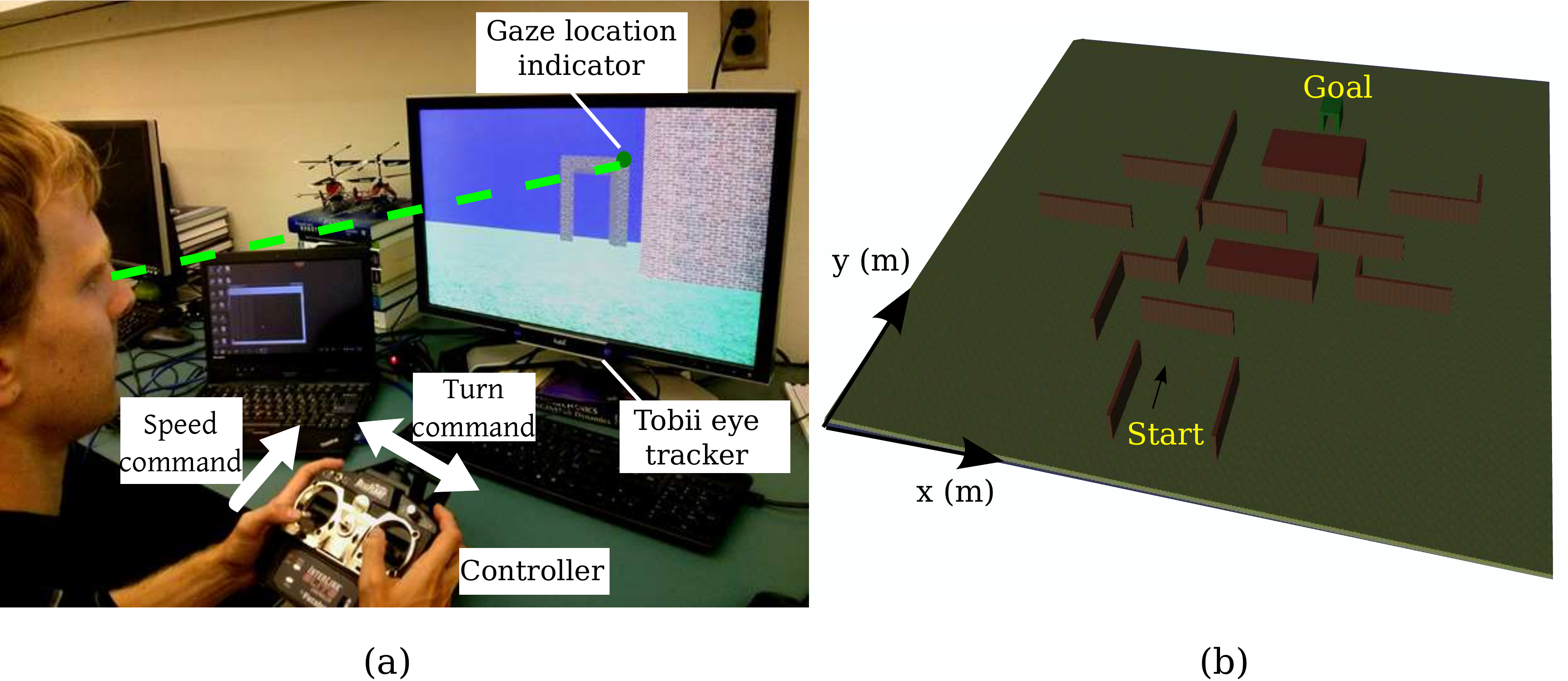} 
\caption{(a) First-person guidance experiment system proposed in~\cite{FeitSMC2015} and (b) Task environment used for human guidance experiments presented in this paper.}
\label{SimSystem}
\end{figure}

Figure~\ref{SimSystem}(b) shows the task environment used for the guidance experiments. The environment is quasi 3-D and made of vertical walls. The experiments in this paper involve only horizontal (planar) motion. Eight subjects participated in the experiments. The task objective was to find fastest (minimum-time) routes between pre-specified start and goal locations as shown in  Fig.~\ref{SimSystem}(b). Before the experiment, the subjects had no knowledge of the environment layout and the goal was described to them as an archway (visually distinguishable from obstacles/walls) situated north of their start orientation. Subjects performed multiple runs from the same start location. At the end of each run, flight-time was displayed on the monitor as a feedback about their performance. Each subject was instructed to try at least 20 runs or as many runs as he/she required to explore the environment in order to find the fastest route. At the end of the experiment, each subject was asked which route was the best (fastest).    

\begin{figure}[htbp]
\centering
\includegraphics[scale=0.45]{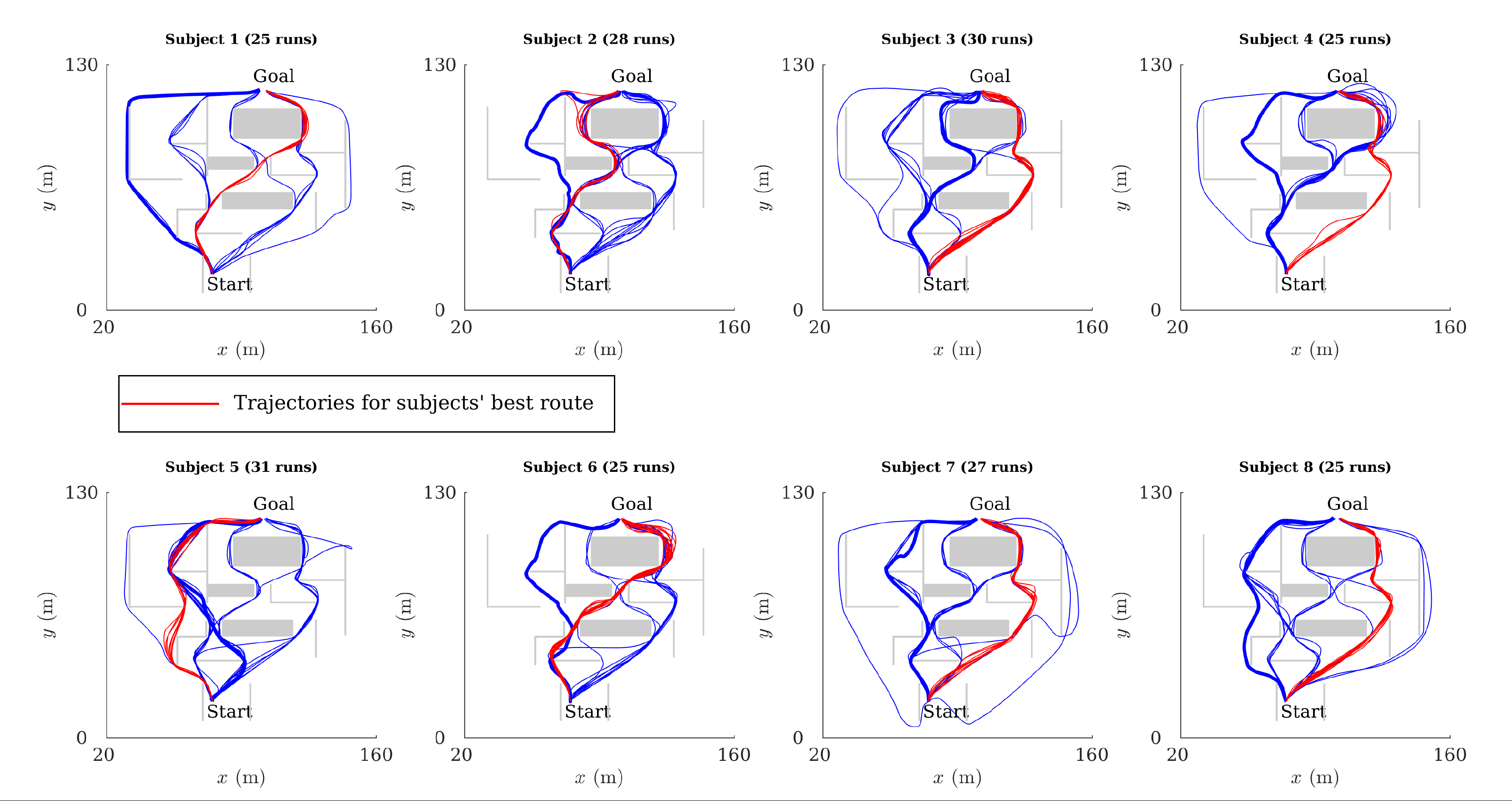} 
\caption{Trajectories for all runs for subjects 1 to 8.}
\label{Trajs}
\end{figure}

\begin{figure}[htbp]
\centering
\includegraphics[scale=0.5]{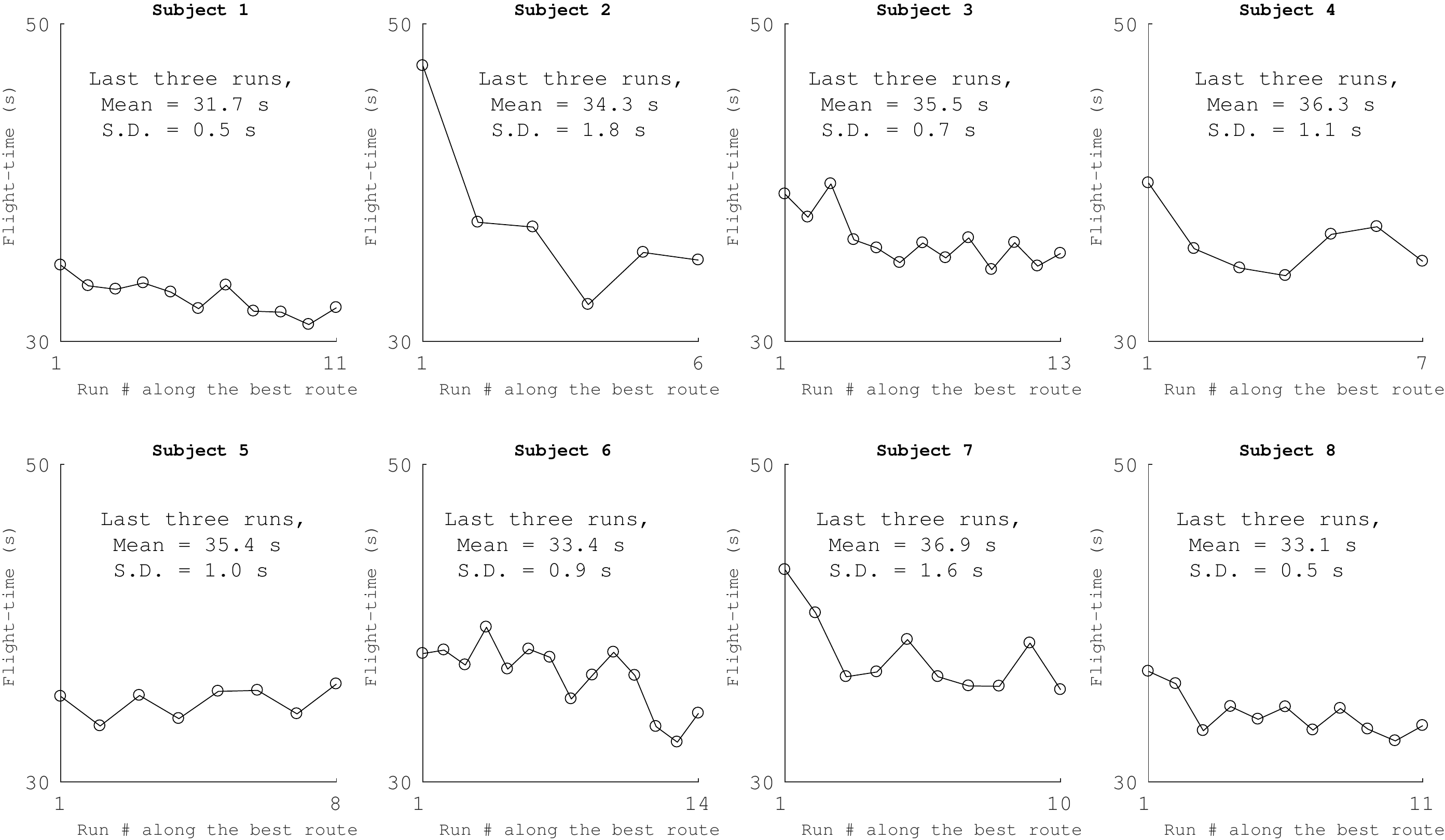} 
\caption{Flight-times for runs on best routes for subjects 1 to 8. S.D. is the standard deviation.}
\label{FlightTimeBest}
\end{figure}

Figure~\ref{Trajs} shows trajectories for all runs for subjects 1 through 8. For each subject, trajectories on his/her best route are shown in red. Figure~\ref{FlightTimeBest} shows the flight-times for runs on the best route for each subject. Subject 1 achieved the best overall flight-time of 31.0 $s$. 

\subsection{Preceding Work}

This section briefly reviews the concepts of spatial value function (SVF), interaction patterns (IPs), and hierarchical model of human pilots' guidance and perceptual behavior. 

\subsubsection{Spatial Value Function (SVF)}

For a trajectory optimization problem in which a vehicle has to reach a specified goal state $\mathbf{x}_g$ from a start state, spatial value function (SVF)~\cite{mettler2010agile} describes optimal cost-to-go (CTG) and velocity vector field (VVF) over a geographical space. 

\subsubsection{Spatial Structures (Patterns) in SVF}

Kong and Mettler~\cite{KongCDC2009} described structural features (subgoals, repelling and attracting manifolds) in the SVF. They investigated these elements using a toy example based on the opitmal solution for a Dubins vehicle that has to reach a goal in an obstacle field. Subgoals partition a task space such that optimal solution in each partition converges to a subgoal. A common boundary of two space partitions is defined as either repelling or attracting manifold. Velocities converge and diverge along attracting and repelling manifolds, respectively. These features make it possible to abstract the solution. The entire solution, i.e., SVF, can be described as a directed graph of subgoals. The subgoal graph representation of task space accounts for both the vehicle dynamics and environment. Trajectory to the goal from any location in the task space can be represented by a subgoal sequence. 

\subsubsection{Human SVF}

Spatial value function (SVF) describes spatial guidance behavior associated with an optimal guidance policy (e.g., cost and velocity maps over geographical space). Mettler and Kong~\cite{mettler2013mapping} showed that the guidance behavior of a trained operator can be described as SVF. They described a method to extract SVF maps from experimental trajectories in a goal interception task. The extracted SVF maps were compared with an optimal policy based on a mass-point model. The results in~\cite{mettler2013mapping} showed that guidance behavior of a trailed pilot was sufficiently stationary in time, and continuous and consistent over the space. Therefore, the concept of SVF is a valid tool for the analysis of human guidance behavior. 
Kong and Mettler~\cite{KongHMSIEEE2013} subsequently extended the analysis to investigate the organization of guidance behavior over large task environments with obstacles. They suggested that humans exploit invariants in the dynamic interactions with the environment to mitigate complexity, which is discussed next. 

\subsubsection{Interaction Patterns: Human Pilot}

The patterns described in Dubins solution space~\cite{KongCDC2009} are a result of interaction between vehicle dynamics and environment. In human-piloted guidance tasks, human operators interact with the task environment through their control, guidance, and perceptual mechanisms. To account for human interactions with the task elements, Kong and Mettler~\cite{KongHMSIEEE2013} used the concept of closed-loop agent-environment dynamics~\cite{Warren2006}. 
The authors showed patterns such as subgoals and guidance primitives in interactions of human pilot's control and guidance mechanisms with the task environment. These patterns are described as interaction patterns (IPs). Pilots use IPs to organize guidance and planning behavior in spatial tasks. A mathematical formulation for IPs is given in Section~\ref{sec:analysis}.


\subsubsection{Functional Model of Human Guidance}

Mettler et. al~\cite{MettlerFrontiers2015} presented a hierarchical multi-loop model of human guidance behavior in spatial control and planning tasks. The model uses IPs (e.g., subgoals and guidance primitives) as organizational units of human spatial behavior. The hierarchical model delineates planning, perception, and control. 
At the highest-level, i.e. planning, a human pilot decomposes the global task into subtasks as a sequence of subgoals. To navigate between subgoals, the pilot deploys a series of guidance primitives that combine extraction of information from immediate environment and control behavior for stereotypical environment conditions.
Thus IPs are considered as units of organization of guidance behavior in a task space. 
Therefore, pilots have to learn IPs when they navigate in unknown environments. IPs can be modeled as units of learning for planning and guidance in unknown environments. 

Mettler et. al~\cite{MettlerFrontiers2015} also presented a hierarchical model of perceptual behavior that models visual attention as a function of three levels (planning, perceptual guidance, and tracking and pursuit) in the hierarchical guidance model.
Andersh et. al~\cite{Andersh2014} tested the hypothesis based on the functional model. The authors investigated visuo-motor control in a remote-control goal-interception task. The analysis showed that pilots' gaze follow the vehicle. 
In between, pilots use saccades to rapidly switch gaze to the goal location and fixate gaze at the goal for a small duration. The smooth pursuit, i.e., gaze following the vehicle, and saccades provide estimates of vehicle velocity and motion gap to the goal location, respectively. 

\subsection{Research Questions}

The general goal of the present work is to apply the functional understanding described above to model the cognitive functions that facilitate environment learning in spatial guidance tasks in humans. Downs and Stea~\cite{Downs1973} gave a formal definition of cognitive mapping: ``Cognitive mapping is a process composed of a series of psychological transformations by which an individual acquires, codes, stores, recalls, and decodes information about the relative locations and attributes of phenomena in his everyday spatial environment." Following the definition in~\cite{Downs1973}, this paper formulates specific questions for environment learning in spatial guidance tasks in humans. The questions are: 1) what information is extracted from interactions with the environment?, 2) what is the memory structure for coding and storing the information?, and 3) how the information is represented to support planning and decision-making?


\subsection{Hypothesis}
\label{subsec:hypothesis}


The interaction patterns (IPs)~\cite{KongHMSIEEE2013,MettlerFrontiers2015} are the main elements needed to describe the behavior, and therefore can be used to abstract the task environment as a graph network of subgoals. The hypothesis for flight tasks in unknown environments is that IPs  serve as basic elements of the memory structure used for learning and representing the task knowledge. 

A skilled pilot learns guidance primitives that represent trajectory maneuvers optimized for interactions with the environment. For the guidance primitive of the skilled pilot, perceptual and control policies are coupled directing the pilot's visual attention, and determine task-relevant features of the environment. The formation of guidance primitives are required for the learning of the optimal subgoals and their network. A subgoal graph representation of the task enables the pilot to focus on the layout of the global plan as a sequence of subgoals and implement control following the guidance policies associated with the IPs.    


\subsection{Paper Outline}

The rest of the paper is organized as follows. Section~\ref{sec:background} presents the research background in human spatial navigation and decision-making. Section~\ref{sec:formulation} presents a formulation of guidance task, interaction patterns, and agent-environment system. Section~\ref{sec:analysis} presents the framework used to analyze human data. Section~\ref{sec:results} presents an analysis of human data for planning, learning, and control behavior. 
 

\section{Background}
\label{sec:background}

This section gives a brief overview of past research in humans/animals spatial memory, representation, navigation and wayfinding, and decision-making. 
 
\subsection{Spatial Memory and Representation}

Humans and animals walk from one site to another in everyday life and it is rarely aimless~\cite{Rosenbaum2009}. A dominant hypothesis has been the theory of spatial memory, which has been investigated by studying rats' movement in mazes (e.g. \cite{Tolman1948,Olton1976,Olton1979}). However, before the theory of spatial memory, walking was thought to be a reflex chain, i.e., a sequence of stimulus-response mechanisms, but the results from experiments with rats in mazes argued against the reflex chain theory~\cite{Tolman1948}. 

\subsubsection{Cognitive Map}
 
Tolman~\cite{Tolman1948} proposed that a rat builds a mental (cognitive) map of the environment (maze) describing routes, paths, and environmental relationships. He proposed that rats use the cognitive map to determine (select) which responses will be released when bombarded by various stimuli in navigating a maze, rather than responding based on stimulus-response relationships. 

\subsubsection{Route vs. Survey Maps} 

Cognitive (mental) maps take two primary forms~\cite{Tolman1948,Rosenbaum2009}: strip-like (route) and comprehensive (survey) maps. A route map encodes a specific path as a series of locations and turns. Such a map is not flexible to changes in the original environment or the start position. The survey map encodes relative positions of landmarks in an environment, and is more reliable than a route map for travelling between any two points in the environment.

Studies on the theory of spatial memory (e.g.,~\cite{Tolman1948,Olton1976,Olton1979,Norman1981}) show that survey and route maps are selected based on the specific application. For example, if one takes a particular route regularly, the travelling process becomes automatized and is better explained by route maps~\cite{Olton1979,Norman1981}. On the other hand, survey maps better explain the behavior of rats in maze when a rat can find a food from a new start position~\cite{Tolman1948,Olton1976}. 

\subsubsection{Spatial Representation in Humans' Brain}

Researchers (e.g. \cite{Stevens1978,Thorndyke1981,Hirtle1985,McNamara1986}) have investigated how humans store spatial relationships within locations in an environment. For example, Stevens and Coupe~\cite{Stevens1978} experimented with human subjects. Based on their observations, they presented a model that stores spatial relationships hierarchically and is governed by ``storage-computation trade off". Spatial relationships that are not stored have to be determined by combining the stored spatial relations. Thorndyke~\cite{Thorndyke1981} showed experiments in which human subjects were asked to estimate distances between two points on a map while viewing the map. Based on observations, the author in~\cite{Thorndyke1981} proposed a model that expresses the estimated distance of a route as a linear combination of the true distance and the number of intervening points on the route. A highly cluttered map corresponds to a large number of intervening points. The hypothesis in~\cite{Thorndyke1981} was that a subject visually scans along a route and judges the distance  based on the scan time. At an intervening point, the subject pauses the scan to check if the point is the destination. Therefore, each intervening point takes a non-zero scan time and increases the overall scan time on the route, which increases the resulting distance estimate.    
Hirtle~\cite{Hirtle1985} investigated humans' spatial representations of natural environments that do not have an obvious or well-defined hierarchical structure. The analysis in~\cite{Hirtle1985} supported the view that a mental model of a real-world environment is composed of both spatial and nonspatial (non-Euclidean) information. The nonspatial information is stored in a hierarchical data structure based on subjective quantities such as intuitively pleasing and stability over time.
McNamara~\cite{McNamara1986} tested three classes of theories of the mental representation of spatial relations, which are nonhierarchical, strongly hierarchical (maximizing storage efficiency by storing minimum spatial relations required to represent a layout accurately), and partially hierarchical (storing many spatial relations that can be induced by other stored spatial relations) theories. Human experiments in \cite{McNamara1986} supported partially hierarchical representation of spatial relationships.

Thomson~\cite{Thomson1980} presented a study of ``blind" walking in humans. Subjects were first showed the target for 5 seconds and then they had to walk blindfolded towards the target. The results showed that performance degraded gradually with increasing target distance (beyond 9 meters). They investigated two hypotheses, perceptual and memory limitations, for the performance degradation. More experiments and analysis in~\cite{Thomson1980} suggested that subjects use the spatial memory (like a mental map or image of the environment) to move towards the target rather than a blind motor program to guide themselves. However, the mental map fades with time (specially beyond 8 seconds). 

Several experimental studies as discussed above have supported the concept of cognitive (hierarchical survey) maps. However, there have been other theories of spatial knowledge such as landmark-based, path-integration, etc. For example, Foo et. al~\cite{Foo2005ExperimentPsychology} presented navigation experiments in a virtual environment. The analysis in~\cite{Foo2005ExperimentPsychology} showed that humans rely on visible landmarks while navigating an environment. The authors in~\cite{Foo2005ExperimentPsychology} suggest that humans spatial knowledge not necessarily fall into any single class (cognitive map, route map, path-integration, landmark-based, or etc.). The present paper uses a graph representation of task space, which is a form of cognitive map, based on interaction patterns to investigate human environment learning in guidance and navigation. The framework based on sensory-motor patterns accounts for dynamic interactions between the agent and the environment, whereas the above studies involve either static or quasi-steady interactions and are discrete decision problems.
The framework also investigates how visibility of nodes in the graph representation affects environment learning, which is similar to how much humans rely on visible landmarks.     
  
\subsubsection{Cognitive Robotics}


Cognitive robotics~\cite{Christaller1999} is inspired from human/animal spatial cognition. Jefferies and Yeap~\cite{Jefferies2008} provided a survey on robotic and cognitive approaches for spatial mapping, to motivate cross-fertilisation between the two areas. As stated in the survey, roboticists work on ``sensor problems" and the cognitive researchers focus on ``knowledge problems". The latter is defined as what people remember most when they visit new places and how they organize spatial information to form knowledge of their environment \cite{Jefferies2008}.
To achieve high-level cognitive capabilities for robots, human or animal spatial cognition has been studied and used to model environments. 
For example, Chakravorty and Junkins~\cite{Chakravorty2007} presented an ``intelligent path planning" method in an uncertain environment. The method uses sensors that allow the sensing of the environment non-locally, which is inspired from human vision. Vasudevan et. al~\cite{Vasudevan2007} proposed a hierarchical probabilistic representation of space, based on high-level environment features such as household objects and doors. The goal was to make robots represent an environment in a way that is comprehensible to humans. Manning et. al~\cite{Manning2014} presented a cognitive map-based computational model for wayfinding, which consists of three primary modules: vision (acquire and process visual information), cognitive map (store spatial information from the vision), and route generation (generate a route using the cognitive map). The model uses two parameters: vision and memory. The vision parameter accounts for accuracy of visual information (e.g., scene in peripheral vision is less accurate). The memory parameter accounts for that spatial memory fades with time. The wayfinding model in~\cite{Manning2014} was able to capture a range of behavior from directed route search to random walking.

 

\subsection{Environment Representation}

Humans' spatial navigation capabilities outperform autonomous robots in versatility, robustness, and effectiveness (e.g., success-rate). Spatial navigation in humans have been studied in the past. For example, Chase~\cite{Chase1983} investigated how taxi drivers navigate in large-scale urban environment that can not be perceived from a single vantage point. The author found that drivers use a hierarchical representation of the environment, which validates the theory of cognitive maps. 
Gillner and Mallot~\cite{Gillner1998} studied the effect of local visual information on human environment learning, using movement data from experiments in a virtual maze. The results indicated that humans learn a maze as a view graph, i.e., sequence of local views and movements. Information at a node includes a recognized position, movement decisions, and expected next views for different decisions.
Spiers and Maguire~\cite{Spiers2008} presented a study of taxi drivers, which involves retrospective verbal reporting by drivers and gaze tracking. The method in~\cite{Spiers2008} allows to do a temporal analysis of thoughts and understand cognitive (thinking) processes relevant to wayfinding.   

\subsubsection{Nested Environments} 

Most real-world environments are nested (e.g., a university campus, buildings in the campus, and then laboratories in the building). A theory is that nested environments are represented by a combination of different representations organized in a nested hierarchy~\cite{Stevens1978,McNamara1986,Wang2003} providing efficient structure for cognitive processing. Wang and Brockmole~\cite{Wang2003} presented an experimental study that concluded that humans don't necessarily incorporate newly learned section of an environment into existing spatial knowledge, but switch between spatial representations when they cross specific spatial regions. At switching location, humans update their orientation information based on new spatial representation. 

\subsubsection{Three-Dimensional Spatial Representation} 

Real-world environments are three-dimensional which makes its spatial representation complicated. Jeffery et. al~\cite{Jeffery2013} suggested that 3-D world are not represented by a fully volumetric map, but are represented by a combination of several planar representations that correspond to the plane of locomotion. Representation in the orthogonal plane to the plane of locomotion is based on some non-metric way, different from the representation in the plane of locomotion. The authors suggested that even animals that move freely in 3-D world (e.g, birds) use such quasi-planar representations. 



\subsubsection{Route Selection (Wayfinding)} 

In everyday navigation tasks, humans have to select a route among many possibilities. Researchers have investigated what factors influence route selection in humans. For example, Golledge~\cite{Golledge1995} experimentally investigated what selection criteria, other than traditional ones such as minimum time, humans use to select a route in a map. Some non-traditional criteria are initial heading (direction of perception), number of stops on a route, fewer turns, shortest leg first, aesthetically pleasing routes, etc. Hochmair and Frank~\cite{Hochmair2000}, for instance, showed that humans use a least-angle strategy at intersections, i.e., select most straight lines, for wayfinding-decisions in unknown street networks.
Hartley et. al~\cite{Hartley2003} showed that cognitive processes are different for travelling a new (or less-travelled) route than a well-known (frequently-travelled) route. For well-known routes, sequences of body movements (motor commands) get automated, which requires less perceptual and attentional processing~\cite{Jueptner1997I,Jueptner1997II}. 

\subsubsection{Cues in Wayfinding} 

Darken and Sibert~\cite{Darken1996} presented a study involving humans in virtual space to investigate what cues can aid humans to improve wayfinding performances. Some cues suggested in~\cite{Darken1996} are directions indicators, path restrictions, absolute reference points, etc. Ruddle at. al~\cite{Ruddle1997} showed that if familiar objects are used as landmarks, the wayfinding performance is better. Waller et. al~\cite{Waller2000} showed that for learning their location, humans may rely more on distance information than bearing information of landmarks, and suggested to account for this finding in modeling human place learning.
Kato and Takeuchi~\cite{Kato2003} showed that a good sense of direction aids a human in wayfinding by assisting in either using a global frame of reference (e.g., cardinal directions) or memorizing landmarks and their relative locations.
Kelly et. al~\cite{Kelly2008} studied the effects of environmental geometry on wayfinding performance. The study in~\cite{Kelly2008} showed that visible angular corners help humans estimate spatial orientation. Vilar et. al~\cite{Vilar2014} experimentally showed that horizontal signage prove more helpful than vertical signage in improving wayfinding performance of humans. 

\subsubsection{Asymmetry in Route Choices} 

Bailsenson et. al~\cite{Bailenson1998,Bailenson2000} investigated why subjects choose different routes if start and target locations are switched. The study in~\cite{Bailenson1998,Bailenson2000} found that subjects prefer routes that have longer and straighter initial segments, which are called hill-climbing or initial segment strategy (ISS). Brunyé et. al~\cite{Brunyé2010,Brunyér2015} showed that some humans have preference for routes that are Southbound. A possible explanation for the southern preference is misperceptions of increased elevation in North direction~\cite{Brunyé2010}. Vreeswijk et. al~\cite{Vreeswijk2013} presented a study that shows that when travel times of two routes are within a range, human drivers are biased with one route and are not willing to alter their choice even if the traffic conditions and other factors change.
 
\subsection{Decision-Making}

Route selection in environment learning and spatial navigation involves decision-making, i.e., selecting a route among many possibilities or choosing between exploring new options and exploiting known ones.  
Simon~\cite{Simon1957} described decision-making as ``a search process guided by aspiration levels. An aspiration level is a value of a goal variable which must be reached or surpassed by a satisfactory decision alternative". This section presents a brief overview of various factors in human decision-making.

\subsubsection{Bounded Rationality and Satisficing}

In traditional optimal control, decision-making refers to optimization of an objective function, which is called rational (optimal) behavior. In classical economics, humans were usually modelled as ``economic (rational) man"~\cite{Simon1957}. Simon~\cite{Simon1957} argued against the economic man assumption and introduced the concept of bounded rationality that accounts for the fact that human decision-making is constrained by limits on time, available information, and cognitive processing capacities. He further introduced the concept of satisficing that replaces the goal of maximizing an objective function~\cite{Simon1957}. A possible way of satisficing is to try available alternatives  in a sequential order and stop when an alternative that meets all criteria of an acceptable solution is found~\cite{Simon1957}. 

Simon~\cite{Simon1956,Simon1957} also suggested that humans use the structures in task environment for their decision-making. In the present research, the concept of invariants (patterns) in agent-environment interactions in guidance tasks accounts for structure in the agent's behavior resulting from its interactions with environment. The interaction patterns \cite{KongHMSIEEE2013,MettlerFrontiers2015} provide a way to abstract the search space which in turn can be represented as a graph. This framework is used here to investigate to model human behavior and decision-making in environment learning.

\subsubsection{Information Processing Model and Working Memory}
 
Humans have limits on their memory and information processing, which is a primary reason that humans are in general not optimizers but satisficers. Cowan~\cite{Cowan1988} presented an information processing model for humans, which consists of long-term memory storage, working memory, and focus of attention. 
Due to limited cognitive processing capabilities, humans can recall or remember only a limited amount of information at a time, which is called working memory. It is defined as a subset of long-term memory. 
Both bottom-up (involuntary factors: salient features in the perceived environment) and top-down (voluntary factors: personal beliefs) factors contribute to what information is held in working memory~\cite{Knudsen2007}. The information held in the working memory forms a basis for decision-making. 

To overcome working memory limitations, a hypothesis is that humans use chunking mechanism~\cite{Miller1956}. In chunking, bits of information that have some type of similarity are combined into larger units called chunks. For example, expert players in chess create perceptual chunks of similar sub-configuration of pieces~\cite{Chase1973}.
Another approach to overcoming the limitations in working memory involves pruning decision trees by using heuristics (e.g., Branch and Bound method~\cite{Land60anautomatic}). Huys et. al~\cite{huys2012bonsai} presented a study of human decision-making in a sequential decision-making task. The results in~\cite{huys2012bonsai} showed that humans stop any further evaluation of a sequence if it exceeds a cost value higher than a threshold. 

 

\subsubsection{Economic vs. Perceptual Choices} 

Human decision-making is in general investigated either on a computational or a neural basis~\cite{Summerfield2012}. These approaches are called economic decision making (EDM) and perceptual decision making (PDM), respectively. In EDM, it is investigated how choices are made based on a value of alternatives. In PDM, the investigation focuses on perceptual properties (e.g., saliency) of alternatives. Towal et. al~\cite{Towal2013} presented a study that shows that a combined model of EDM and PDM is more accurate for humans than either model alone. 
   
A number of studies have shown that gaze fixations create a bias in decision-making~\cite{Shimojo2003,Krajbich2010,Krajbich2011,Towal2013,sakellaridi2015cognitive}. For example, Shimojo et. al~\cite{Shimojo2003} modelled gaze bias as a ``cascade effect". According to the cascade effect model, in starting the gaze is evenly distributed between alternatives and it gradually shifts to the option that is eventually selected. Krajbich et. al~\cite{Krajbich2010} showed that the probability of first-seen option being selected increases with the duration of first fixation. The early gaze bias~\cite{Krajbich2010} was observed by Sakellaridi et. al~\cite{sakellaridi2015cognitive} in his study of visual exploration of city maps. In~\cite{sakellaridi2015cognitive}, humans subjects were asked to look at a city map and asked to choose a target (from given choices) to go to from a centre point on the map. The eye fixation analysis in~\cite{sakellaridi2015cognitive} showed that humans shown an early selection bias even from the beginning of a trial.

\section{Mathematical Formulation}
\label{sec:formulation}
This section first presents a mathematical formulation of guidance task using the modeling language associated with the interaction patterns. Next, the formulation is used to model memory structure for representing and learning a guidance task environment. Finally, it presents the agent-environment system and its components.

\subsection{Guidance Task}

In a guidance task, an agent travels from a state $\bf{x} \in$ $\raisebox{1.5pt}{$\chi$}$ $\subseteq \mathbb{R}^n$ to a given goal state ${\bf{x}}_g$, using control $\bf{u} \in$ $\raisebox{0.0pt}{$\mathcal{U}$}$ $\subseteq \mathbb{R}^m$. 
Vehicle dynamics are described by:
\begin{eqnarray}
\dot{\bf{x}} = f(\bf{x},\bf{u}), \\ \nonumber
{\bf{x}}_p \in \raisebox{0.0pt}{$\mathcal{W}$} \subset \raisebox{1.5pt}{$\chi$},
\label{Dynamics}
\end{eqnarray}
where ${\bf{x}}_p$ is spatial position vector and $\raisebox{0.0pt}{$\mathcal{W}$}$ is allowed workspace (e.g., position and orientation). The time to reach the goal is represented by $t_f$. A control trajectory $\overleftarrow{u}$ drives the agent from a start state $\bf{x}$ to the goal state ${\bf{x}}_g$. The corresponding state trajectory is represented by $\overleftarrow{s}$. The set of all feasible trajectories from all start states satisfying constraints $\raisebox{1.5pt}{$\chi$}$ and $\raisebox{0.0pt}{$\mathcal{W}$}$ is represented by $\overleftarrow{S}$, which represents guidance behavior.

An optimal trajectory ($\overleftarrow{u}^*$ and $\overleftarrow{s}^*$) minimizes a cost function $J$ (e.g., time-to-go) as follows: 
\begin{equation}
\Min_{\overleftarrow{u}} \int_{0}^{t_f} J(\bf{x}(t),\bf{u}(t)) dt.
\end{equation}
The set of $\overleftarrow{s}^*$ from all start states is represented by $\overleftarrow{S}^* \subset \overleftarrow{S}$, which represents optimal guidance behavior.
Optimal spatial guidance behavior $\overleftarrow{S}^*_p$ is defined over spatial position vector ${\bf{x}}_p$ space. $\overleftarrow{S}^*_p$ is the set of optimal trajectories from all ${\bf{x}}_p \in \raisebox{0.0pt}{$\mathcal{W}$}$. Spatial value function (SVF) describes optimal
guidance policy (e.g., cost-to-go (CTG) and velocity maps) over geographical space for $\overleftarrow{S}^*_p$.

\subsection{Interaction Patterns}

Kong and Mettler~\cite{KongHMSIEEE2013} described two equivalence relations that are fundamental to the organization of spatial behavior: subgoals ($g$'s) equivalence and the symmetry group guidance primitives ($\pi$'s), in $\overleftarrow{S}^*_p$. These two equivalences provide the elements to formally describe patterns in interactions between agent dynamics and environment. 

A subgoal $g \in \raisebox{1.5pt}{$\chi$}$ is a state that two trajectories $\overleftarrow{s}^*_i$ and $\overleftarrow{s}^*_j$, in $\overleftarrow{S}^*_p$, meet at and then follow a same trajectory to the goal. Trajectories related by a same subgoal $g$ are said to be equivalent, i.e., $\overleftarrow{s}^*_i \sim_S \overleftarrow{s}^*_j$. 
Subgoals divide the task space $\raisebox{0.0pt}{$\mathcal{W}$}$ into partitions $\raisebox{0.0pt}{$\mathcal{W}$}_i$'s such that trajectories from all ${\bf{x}}_p \in \raisebox{0.0pt}{$\mathcal{W}$}_i$ converge to the same subgoal $g_i$. Therefore, trajectory $\overleftarrow{s}^*$ from a point can be represented as a sequence of subgoal states.
 
A trajectory segment is a continuous portion from a trajectory $\overleftarrow{s}^*_i$. If two trajectory segments $\pi_i$ and $\pi_j$ are equivalent after a rigid-body transformation (translation and rotation), the segments are related to same guidance primitive, i.e., $\pi_i \sim_G \pi_j$. The guidance primitive library $\Pi$ is as follows:
\begin{equation}
\Pi = \{ \pi_1, \pi_2, ... \}
\end{equation} 
A trajectory $\overleftarrow{s}^*$ can be represented as a string of guidance primitives.   

\subsection{Subgoal Graph}
The optimal guidance solution over spatial position vector, which is $\overleftarrow{S}^*_p$, can be abstracted as a directed graph of subgoals represented by $G$ as follows:
\begin{eqnarray}
G = [g_0 \ g_1 \ g_2 \ .. \ g_k \ .. \ g_N], \\ \nonumber
(g_k)_c = g_i \ \& \ CTG_i < CTG_k, 
\end{eqnarray}
where $N$ is the number of subgoals. The goal is represented by $g_0 = {\bf{x}}_g$. $CTG_k$ is cost-to-go to the goal state ($g_0$) from subgoal $g_k$. $CTG_0$ is zero. Each subgoal (other than goal) $g_k$ has one child subgoal $g_i$, i.e., there is a directed edge in the graph from node $g_k$ to $g_i$. Graph edges are represented by a connection matrix $Q$ as follows:
\begin{eqnarray}
Q = {[Q_{ki}]_{(N+1) \times (N+1)}}, k \in [0 \ .. \ N], i \in [0 \ .. \ N]; \\ \nonumber
Q_{kk} = 0 \ \forall \ k \in [0 \ .. \ N]; \\ \nonumber 
Q_{0i} = 0 \ \forall \ i \in [0 \ .. \ N]; \\ \nonumber 
\forall \ k \in [1 \ .. \ N], \ \exists! \ i \ (Q_{ki} = 1, \ Q_{kj}=0 \ \forall \ j \neq i).
\end{eqnarray}
The matrix element $Q_{ki}$ is 1 only if $g_i$ is the child subgoal of $g_k$, otherwise $Q_{ki}$ is 0.

State vector $\bf{x}$ is position ${\bf{x}}_p$ and dynamic (e.g., velocity and higher derivatives) state ${\bf{x}}_v$.
In presented experiments, position vector is $[x \ y]$ and dynamic state is velocity $[v \ \psi]$ where $v$ and $\psi$ are velocity magnitude and direction, respectively. A subgoal $g_k$ is ${\bf{x}}_{g_k} = [x_{g_k}  \ y_{g_k} \ v_{g_k} \ \psi_{g_k}]$. The subgoal position ${{\bf{x}}_p}_{g_k} = [x_{g_k}  \ y_{g_k}]$ coincides with obstacle boundaries (or corners in polygonal obstacle fields)~\cite{KongCDC2009,kong2011foundations}. The subgoal velocity ${{\bf{x}}_v}_{g_k} = [v_{g_k} \ \psi_{g_k}]$ depends on its child subgoal state ${\bf{x}}_{(g_k)_c} = [x_{(g_k)_c}  \ y_{(g_k)_c} \ v_{(g_k)_c} \ \psi_{(g_k)_c}]$ as follows:
\begin{equation}
\Min_{v_{g_k}, \ \psi_{g_k}} \int_{{{\bf{x}}_p}_{g_k}}^{{\bf{x}}_{(g_k)_c}} J(\bf{x}(t),\bf{u}(t)) dt 
\label{subgoal_velocity}
\end{equation}
For a Dubins dynamics (e.g., no acceleration constraint), velocity direction $\psi_{(g_k)_c}$ will coincide with the edges of the visibility graph of subgoal positions, which is as follows:
\begin{equation}
\psi_{(g_k)_c} = {\tan}^{-1} \left[ {y_{(g_k)_c} - y_{g_k}} \over {x_{(g_k)_c} - x_{g_k}} \right].
\end{equation}

Formulation for subgoal velocity (Eq.~\ref{subgoal_velocity}) is a two-point boundary value optimization, which is usually solved using numerical techniques. With a finite library of guidance primitives ($\Pi=\{ \pi_i \}$) as units for motion planning, the optimization problem in Eq.~\ref{subgoal_velocity} can be converted to finding the optimal sequence of guidance primitives to transition between subgoals. 

\subsection{Learning} 
\subsubsection{Subgoal Graph}
In unknown environments, the agent has to learn the subgoal graph $G$. The task environment in presented experiments is made of polygonal obstacles, and therefore subgoal positions are assumed to be associated with obstacle corners~\cite{KongCDC2009,kong2011foundations}. The connection matrix for the agent is a probability distribution as follows:
\begin{eqnarray}
\sum_{i=0}^{i=N} Q_{ki} = 1,
\end{eqnarray}
where $Q_{ki}$ is the probability that $g_i$ is the child subgoal of $g_k$. An approximation of a priori $Q_{ki}$ is as follows:
\begin{equation}
Q_{ki} = \begin{cases} 1/M, & \mbox{if } \mbox{ $V(k,i)$=1} \\ 0, & \mbox{if } \mbox{ $V(k,i)$=0}, \end{cases}
\end{equation}
where $M$ is the number of subgoals that are connected with $g_k$ in visibility graph $V$. With environment learning, the child subgoal is learned, i.e., $Q_{ki}$ shifts to 1 for a particular $i$ and zero for all others. 

\subsubsection{Guidance Primitive Library}

$\Pi_F$ is the set of trajectory-segments associated with guidance primitives $\pi$, which satisfy the vehicle dynamics $f$ and state constraints $\raisebox{1.5pt}{$\chi$}$. Two same trajectories are represented by a same $\pi$. $\Pi_W \subseteq \Pi_F$ is the set of trajectory-segments that are formed through repeated interactions with task environment $\raisebox{0.0pt}{$\mathcal{W}$}$. $\Pi^*_W \subseteq \Pi_W$ includes trajectory-segments that are optimal for a cost function (e.g., time).  

Before the environment is learned, the agent's library $\Pi$ can be assumed to be:
\begin{equation}
\Pi \subset \Pi_F.
\end{equation}
Following the agent's initial interactions with the task environment, $\Pi$ becomes:
\begin{equation}
\Pi \subset \Pi_W.
\end{equation}
Overtime the agent learns optimal guidance policies or primitives, $\Pi$ becomes:
\begin{equation}
\Pi \subset \Pi^*_W.
\end{equation}
When the task environment is learned, the library consists of trajectory segments that are specific for the task environment. 

\subsubsection{Learning Assessment}

Learning therefore can be assessed by changes in $\Pi$. Environment learning can be measured by two quantities: 1) number of dominant guidance primitives in $\Pi$, 2) consistency of each guidance primitive $\pi$. 

Hypothesis is that a subject uses a series of guidance primitives to travel between subgoals. The computational complexity increases with the number of guidance primitives in the library $\Pi$. A proficient subject is supposed to converge to a small set of guidance primitives.   

A skilled operator is supposed to have reliable control over vehicle dynamics and can consistently execute a guidance primitive $\pi_i \in \Pi$. The consistency of a pilot's control is measured by computing the  variance of trajectories that belong to a guidance primitive.  

\subsection{Agent-Environment System}


Figure~\ref{FirstPersonView} shows an example of first-person view of the task environment. 
The agent-environment system has three elements: 1) vehicle dynamics (forward speed $v$ and turnrate $\omega$), 2) human gaze vector $\vec{r}_g$ (distance $r_g$ and angle $\theta_g$ in agent's body frame), and 3) environment cues . 

\subsubsection{Vehicle Dynamics}

\begin{figure}[htbp!]
\centering
\includegraphics[scale=0.6]{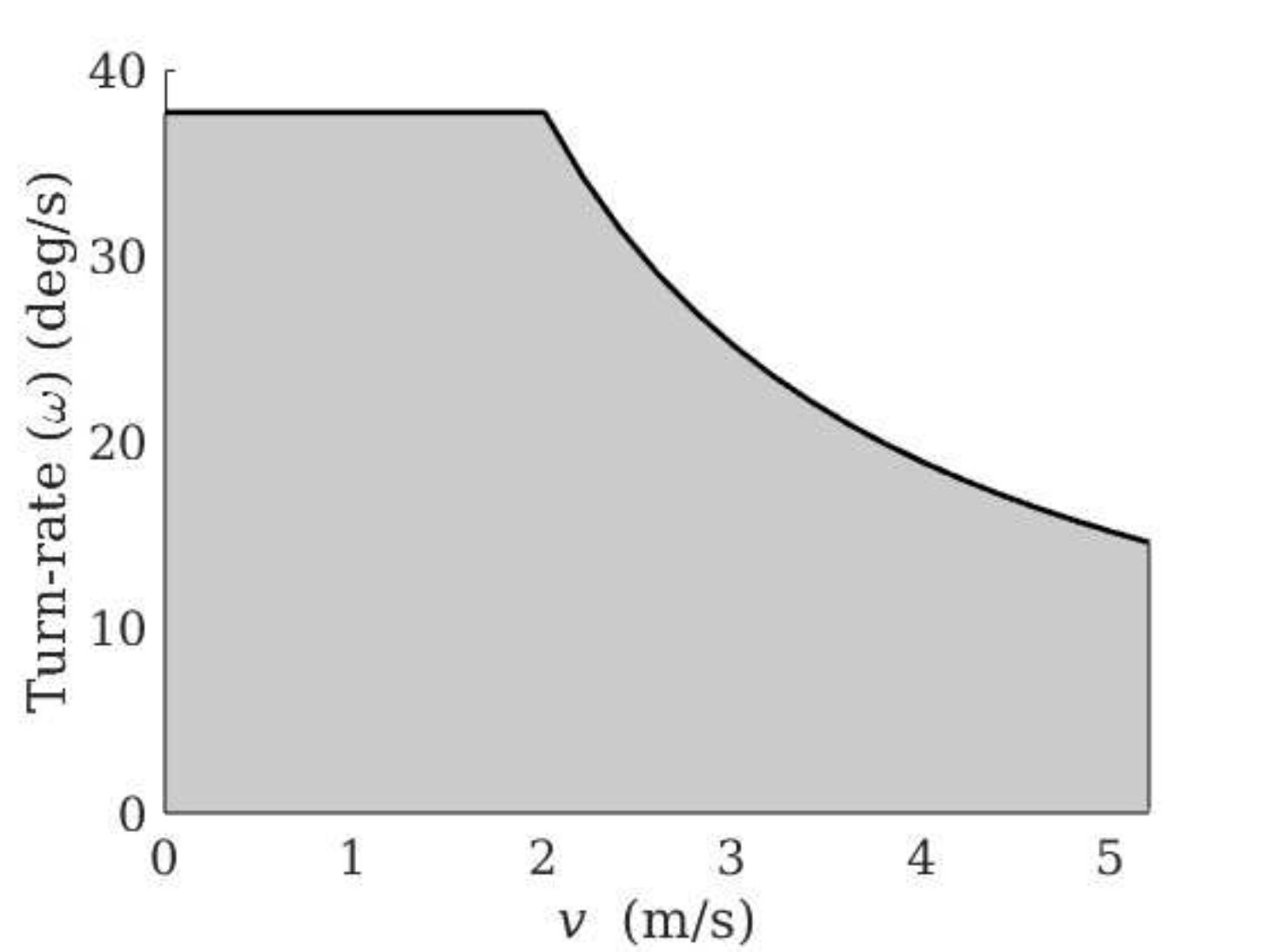} 
\caption{Speed turnrate envelope of the vehicle used in human experiments.}
\label{Speedturnrate_humanexp}
\end{figure}

The forward speed $v$ and turnrate $\omega$ are controlled by longitudinal ($u_{lon}$) and lateral ($u_{lat}$) inputs, respectively. Turnrate is limited based on vehicle speed. Vehicle dynamics model is as follows:
\begin{eqnarray}
\begin{bmatrix}
    \dot{x} \\
    \dot{y} \\
    \dot{\psi} 
\end{bmatrix}
&=&
\begin{bmatrix}
    v \cos{\psi} \\
    v \sin{\psi} \\
    \text{min}(u_{lat}/v,{\omega}_{max})
\end{bmatrix} \\ \nonumber
\dot{v} &=& k_{acc} u_{lon} - k_{drag} v,
\end{eqnarray}
where $\omega_{max}$ is the maximum allowed turn-rate. $k_{acc}$ and $k_{drag}$ are acceleration and drag coefficients, respectively. $v_{max}$ is the maximum speed. In experiments, the values are set as the following:
\begin{equation}
v_{max} = 5.20 \ {m/s}; \ \omega_{max} = 0.65 \ {rad/s}; \ k_{acc} = 7.50 \ {1/s}; \ k_{drag} = 0.88 \ {1/s}.
\end{equation}
Data sampling time $\Delta t$ is 0.02 $s$. Commands $u_{lon}$ and $u_{lat}$ are constrained as follows:
\begin{equation}
0 \le u_{lon} \le 0.62 \ {1/s} \ ; -0.75 \ {m/s^2} \le u_{lat} \le 0.75 \ {m/s^2}. 
\end{equation} 
Figure~\ref{Speedturnrate_humanexp} shows the speed turnrate envelope for the vehicle used in human experiments. 

\subsubsection{Environment Cues}

A cue is a signal used to gain information about some property of the surrounding world. Cues can be visual, auditory, or different sensory types. Visual cues are dominant for humans. In this research, the simulated task environment is made of polygonal obstacles that have two primary features, edges and corners. To keep the environmental cues simple enough for analysis, the simulated environment is presented otherwise homogeneously, i.e., uniform colors for walls and ground, and no other landmarks. Even in an environment composed of polygonal walls, many types of cues are possible, such as a gap between two walls, a point on the edge, lateral or longitudinal distance from the walls. A human subject may use any of these cues to assess his/her state relative to the environment, maintain a safe distance from obstacles, or perceptual guidance (e.g., Tau guidance). For global planning, however, a subject activates a subgoal and approaches the subgoal. Obstacle corners serve as candidates for subgoals. Therefore, the corners or endpoints of the known/learned obstacle boundary can be described as global navigation cues (GNCs) that aid global path planning and navigation. 



\begin{figure}[htbp]
\centering
\includegraphics[scale=0.64]{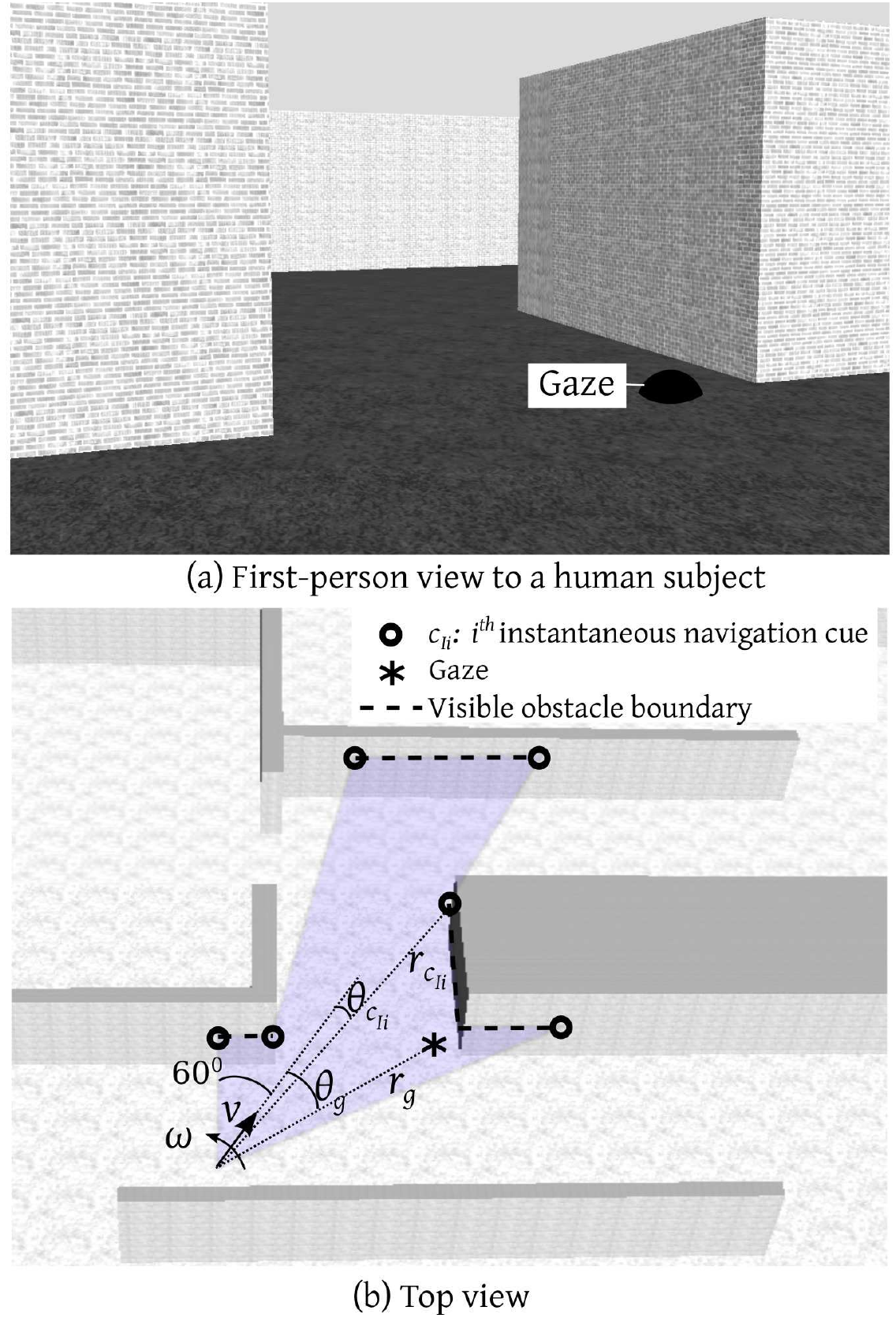} 
\caption{Agent-environment system measurements.}
\label{FirstPersonView}
\end{figure}

An instantaneous navigation cue (INC) is an end point on the visible obstacle boundary as shown in Figure~\ref{FirstPersonView}(b). An INC is represented by $c_I = [r_{c_I} \ \theta_{c_I}]$ where $r_{c_I}$ and $\theta_{c_I}$ are cue distance and bearing angle in agent's body frame. 

\begin{figure}[htbp]
\centering
\includegraphics[scale=0.8]{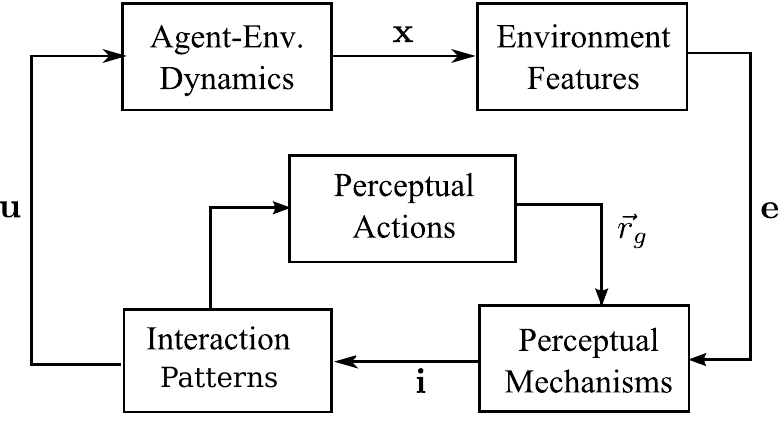} 
\caption{Agent-environment dynamics. $\bf{x}$, $\bf{u}$, $\bf{e}$, and $\bf{i}$ represent agent state, control, environment state, and information extracted from environment cues, respectively. $\vec{r_{g}}$ is the gaze position vector in agent body frame.}
\label{AgentEnvironmentdynamics}
\end{figure}

\subsubsection{Agent-Environment Dynamics}

Warren (2006) described closed-loop agent-environment dynamics (Fig.~\ref{AgentEnvironmentdynamics}) using the following formulation:
\begin{equation}
\dot{\mathbf{x}} = f(\mathbf{x},k(\mathbf{x},h(g(\mathbf{x})))),
\label{closedloop}
\end{equation}. 
The agent is considered to be embedded in the environment. In the closed-loop model Eq.~\ref{closedloop},  
$g(.)$ describes how the agent state affects the environment state $\bf{e}$. For example, environment state can be defined by relative position and orientation of obstacles and navigation cues $c_{I}$'s (subgoal heuristics), which depend on the agent's current state. Next, perceptual processes $\bf{i}=h(\bf{e})$ use environment cues to extract information $\bf{i}$. For example, relative bearing of obstacles can be used to estimate motion gap for perceptual guidance. Navigation cues are used for subgoal selection (decision-making) using a priori known and learned knowledge about task structure (subgoal graph). Next, the agent applies control $\bf{u}=k(\bf{i})$ based on a guidance primitive $\pi_k$ from its guidance primitive library $\Pi$, and moves gaze in a coupling with $\pi_k$. 


\section{Analysis Framework}
\label{sec:analysis}

This section first uses a Dubins vehicle to illustrate the subgoal graph for the task environment used in human guidance experiments. the section applies the subgoal graph model presented in Section~\ref{sec:formulation} for human data processing. Third, it presents an optimal (benchmark) decision-making model to evaluate human decision-making. Fourth, the section presents an exploration metric. Finally, a clustering method to extract guidance primitives is presented.

\subsection{Benchmark Subgoal Graph}

\begin{figure}[htbp]
\centering
\includegraphics[scale=0.42]{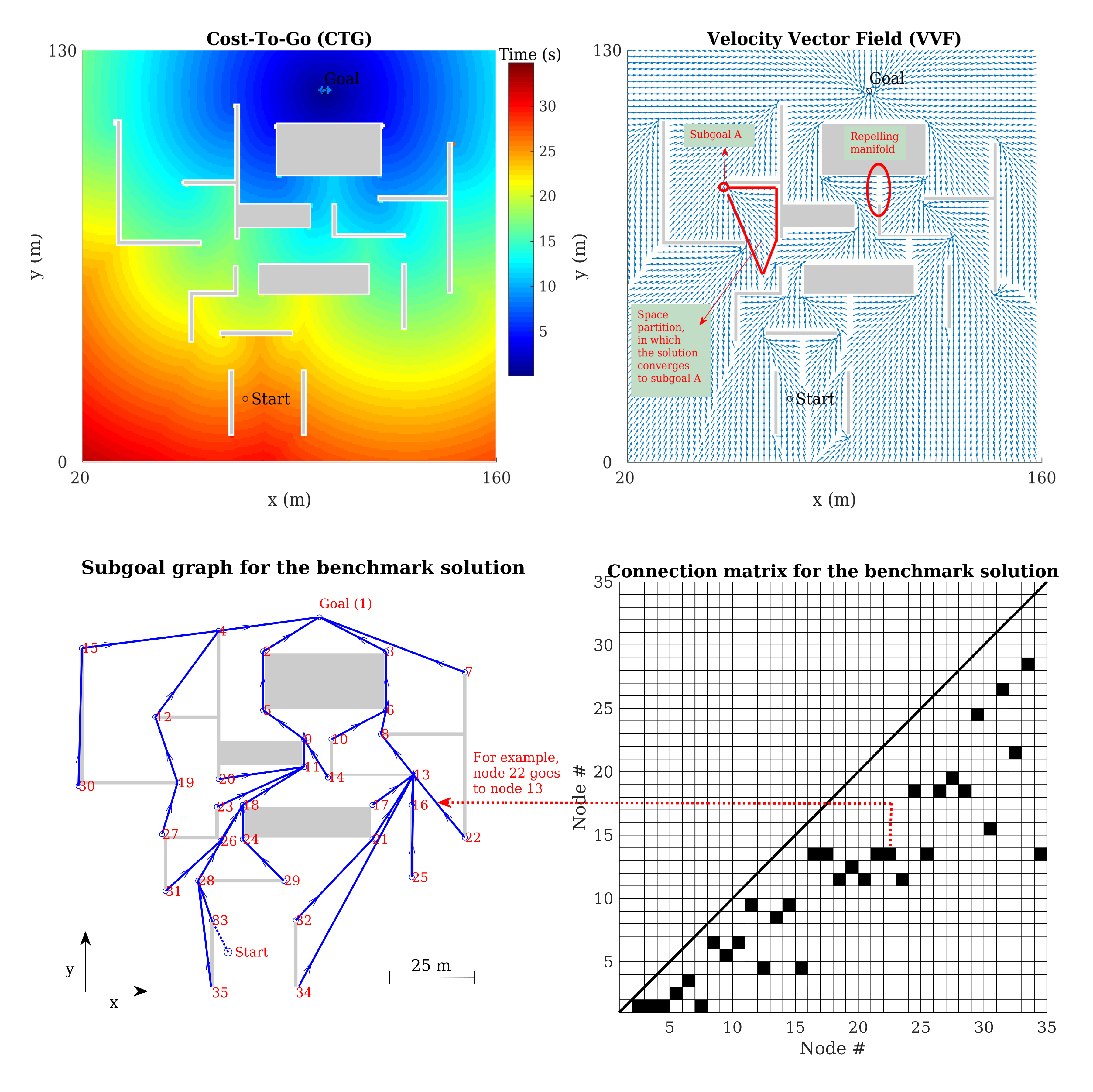} 
\caption{Benchmark solution: Dubins optimal solution, subgoal graph, and connection matrix.}
\label{BenchmarkExample}
\end{figure}

The paper uses the time-optimal solution for a Dubins vehicle (speed and turning radius of $v_{max}$ = 5.2 $m/s$ and 1 $m$, respectively) as a benchmark solution for the task environment shown in Fig.~\ref{SimSystem}(b). Figure~\ref{BenchmarkExample} shows the optimal cost(time)-to-go and velocity vector field for the benchmark solution. The structures such as subgoals and repelling manifold, as described in~\cite{KongCDC2009}, can be seen in the velocity map in Fig.~\ref{BenchmarkExample}. For the optimal Dubins solution, subgoal locations coincide with obstacle corners.

Figure~\ref{BenchmarkExample} also shows the subgoal graph representation, based on the benchmark solution in Fig.~\ref{BenchmarkExample}, for the task environment. A subgoal graph is a directed graph as shown in Fig.~\ref{BenchmarkExample}. Terms `subgoal' and `node' are used interchangeably in this paper.
The solution from each point in free space goes to a subgoal and then it follows a sequence of subgoals (nodes). For example, the subgoal sequence from the start location is start$\rightarrow33\rightarrow28\rightarrow26\rightarrow18\rightarrow11\rightarrow9\rightarrow5\rightarrow2\rightarrow1$(goal) .

An optimal subgoal graph satisfies the dynamic programming formulation (Algorithm 1 in \cite{VermaSMC2015}) as follows:
\begin{equation}\label{eq:CTGij}
CTG_{k} = \min_{{i} \in [0 \ .. \ N] \setminus k} \left( DC_{ki} + CTG_{i} \right) \forall k \in [1 \ .. \ N], 
\end{equation} 
where $DC_{ki}$ is the incremental cost-to-go from subgoal $g_k$ to subgoal $g_i$. 
$DC$ is $(N+1) \times (N+1)$ matrix. A transition from $g_k$ to $g_i$ is allowed only if the optimal trajectory from $g_k$ to $g_i$ in the absence of obstacles is collision-free in the presence of obstacles. 

\subsection{Human Data Processing}
$N$ Nodes, $CTG$, $DC$, and $Q$ are a priori unknown to subjects. Subjects arguably learn these quantities over successive runs. This section describes how to extract learned cost-to-go and node connectivity information from human data.   

A characteristic of a time-optimal trajectory is that it passes close to obstacle corners. This attribute can also be seen in human trajectories (see Fig.~\ref{Trajs}). This characteristic of time-optimal solutions enable the presentation of a human trajectory as a sequence of subgoals $[{k_1} \ {k_2} \ .. \ {k_i} \ {k_{i+1}} \ .. \ 0]$, where $k_i$ is the index of subgoal $g_{k_i}$ in the benchmark subgoal graph. Human cost-to-go at a subgoal $g_{k_i}$ is represented by $CTG'_{k_i}$ and is extracted from a trajectory as follows:
\begin{equation}
CTG'_{k_i} = t_0 - t_{k_i},
\end{equation}
where $t_0$ and $t_{k_i}$ are times at goal and at trajectory point closest to the subgoal $g_{k_i}$'s position, respectively. $CTG'_{k_i}$ from a run is tracked in a list ${CTG'_{k_i}}_{list}$.
For a human subject, $Q'$ is initiated as a zero matrix. In each run, $Q'$ is updated as follows:
\begin{equation}
Q'_{k_i k_{i+1}}=Q'_{k_i k_{i+1}} + 1.
\end{equation} 
Incremental cost between consecutive subgoals in human trajectory is extracted as follows:
\begin{equation}
DC'_{k_i k_{i+1}}=t_{k_{i+1}} - t_{k_i}.
\end{equation}
$DC'_{k_i k_{i+1}}$ from each run is stored in a list ${DC'_{k_i k_{i+1}}}_{list}$.

In the presented framework, human environment knowledge is represented by cost-to-go at nodes ($CTG'_{k_{list}}$), travelling cost from one node to another ($DC'_{{ki}_{list}}$), and number of times a segment from one node to another has been travelled ($Q'_{ki}$). The following are definitions regarding human knowledge about the environment, which will be used to present a decision-making rule later in this section: 

\begin{figure}[htbp]
\centering
\includegraphics[scale=0.65]{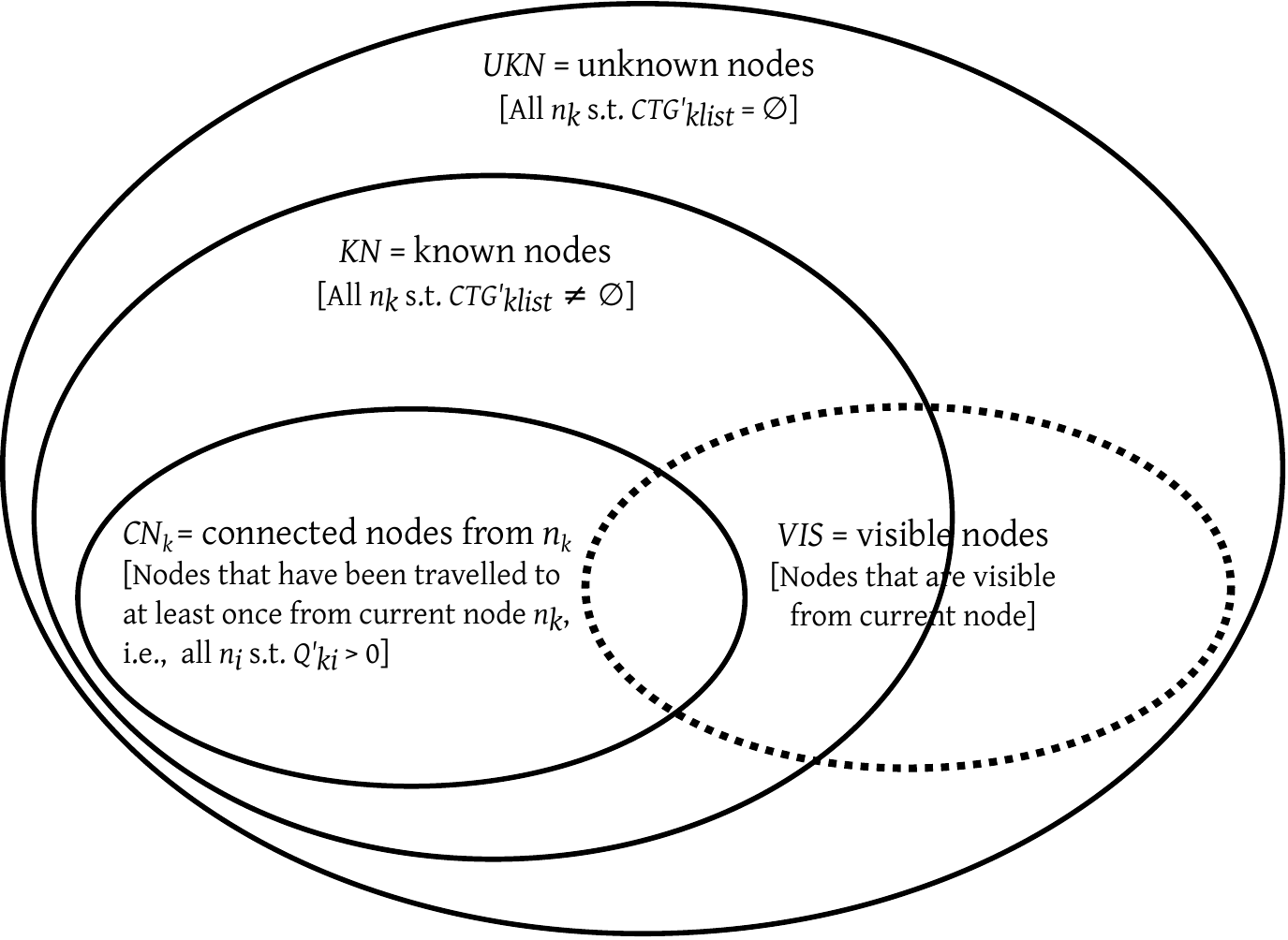} 
\caption{Known, unknown, connected, and visible nodes.}
\label{UnknownKnownNodes}
\end{figure}

\begin{mydef}
Unknown Nodes ($UKN$) is the set of all nodes that have never been visited, and is presented as follows:
\begin{equation}
UKN = \lbrace k \in [1 \ .. \ N]: CTG'_{k_{list}} = \emptyset \rbrace
\end{equation}
\end{mydef}

\begin{mydef}
Known Nodes ($KN$) is the set of all nodes that have been visited at least once, and is presented as follows:
\begin{equation}
KN = \lbrace k \in [1 \ .. \ N]: CTG'_{k_{list}} \neq \emptyset \rbrace 
\end{equation}
\end{mydef}

\begin{mydef}
Connected Nodes at a node $n_k$, represented by $CN_k$, is the set of all nodes that have been travelled to from the node $n_k$, and is presented as follows:
\begin{equation}
CN_k = \lbrace i \in [0 \ .. \ N]: Q'_{ki} > 0 \rbrace 
\end{equation}
\end{mydef}

\subsubsection{Visible Nodes}

In the presented experiments, subjects have a limited field of view (60$\degree$) which is expected to affect their exploratory behavior and choices of routes. A subject has to decide which node to go to after the current node. To study the effect of visibility on decision-making, the set of nodes that are visible from the current node $n_{curr}$ is tracked in $VIS$. $t^*$ is the time at which trajectory is closest to $n_{curr}$. This paper uses a time window $t_w$ around $t=t^*$ to evaluate all nodes visible at any instant from $t=t^*-t_w/2$ to $t=t^*+t_w/2$. They are then stored in $VIS$. If $t_w$ is too big, there are too many overlaps and variables are confounded. A very small $t_w$ is unrealistic from human attention span standpoint. Therefore, it is necessary to identify $t_w$ that explains human behavior and decision-making at nodes. At this point, $t_w$ is set to 1 $s$. 

\subsection{Decision-Making Model}
This section presents Dijkstra's algorithm for shortest path search in human-learned subgoal graph. The algorithm gives a decision-rule to evaluate human decision-making in navigation tasks. 

\subsubsection{Decision Cases at a Node}

At a node, there are two primary types of behavior possible (see table~\ref{tb:choice}): exploration or exploitation, which correspond to trying a new solution or repeating a known solution, respectively. In exploration mode, a subject at a current node $n_k$ goes to a next node $n_i$ that was never visited from $n_k$ ($Q'_{ki}=0$ or $n_i \not\in CN_k$) in preceding runs. In exploitation mode, the subject goes to a next node $n_i$ that was previously visited from the current node $n_k$ ($Q'_{ki}>0$ or $n_i \in CN_k$) in one or more preceding runs. 

\begin{table*}[ht] 
\begin{center}
\caption{Choice at a node $n_{k}$.}\label{tb:choice}
\begin{tabular}{ |p{2.5cm}|p{11.4cm}|  }
 \hline
 Decision case & Choices\\ \hline \hline
 A) $|CN_k|=0$  & Exploration: go to any node \\ \hline
 B) $|CN_k|=1$  & 1) Exploitation: go to the node $n_i \in CN_k$ \\ 
              & 2) Exploration: go to a new node $n_i \not\in CN_k$ \\ \hline
 C) $|CN_k|>1$  & 1) Exploitation: go to a node $n_i \in CN_k$ 
                                (what is the decision-rule?) \\ 
              & 2) Exploration: go to a new node $n_i \not\in CN_k$ \\  \hline
\end{tabular}
\end{center}
\end{table*}

Table~\ref{tb:choice} shows the three types of decision-making scenarios (called cases A, B, and C) at a current node $n_k$. In case A, there is no connected node ($|CN_k|=0$) from node $n_k$, i.e., there is no node $n_i$ that $Q'_{ki}>0$. In cases B and C, there are only one connected node ($|CN_k|=1$) and two or more connected nodes ($|CN_k|>1$), respectively, from node $n_k$. Frequency of case A reduces and increases for cases B and C as a subject learns the environment over successive runs. 
    
\subsubsection{Decision-Making Model}
 
\begin{figure}[htbp]
\centering
\includegraphics[scale=0.5]{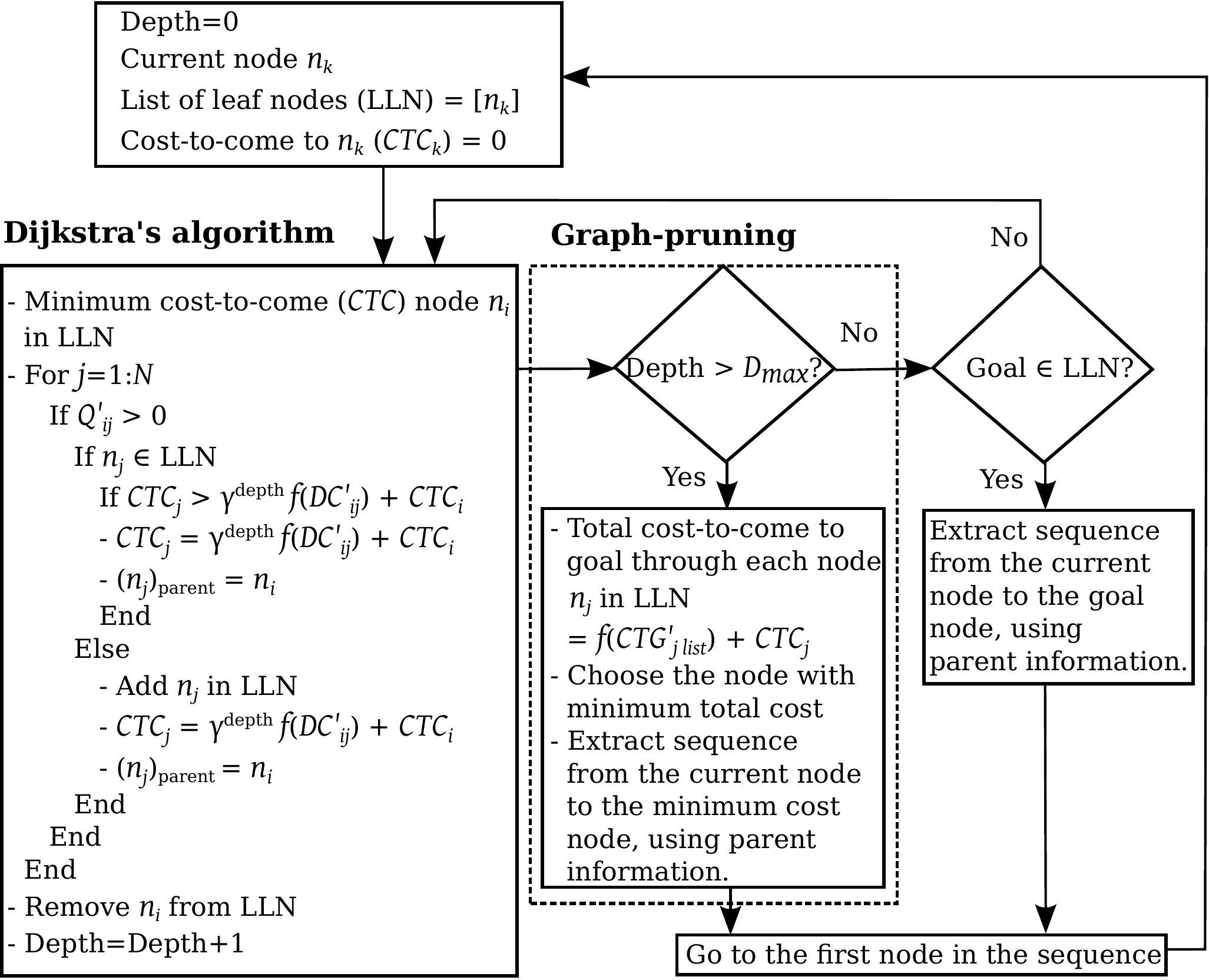} 
\caption{Decision-making model: Dijkstra's algorithm with discount factor $\gamma$ and graph pruning at maximum depth $D_{max}$.}\label{graphsearch}
\end{figure}

Figure~\ref{graphsearch} presents a decision-making model based on the Dijkstra's shortest-path search method proposed in~\cite{Dijkstra1959}. The model is used to select the best node to go in case C (table~\ref{tb:choice}). The decision-making model has two parameters: discount factor ($\gamma$) and maximum depth ($D_{max}$) for graph pruning. 
In a run, the model uses the $CTG'_{k_{list}}$, $DC'_{{ki}_{list}}$, and $Q'$ information extracted from data in preceding runs. At any node, the model uses Dijkstra's algorithm to search for the shortest path to the goal node. The graph is expanded from a node using $Q'$ information. The cost of an edge is given by a function $f(DC'_{{ki}_{list}})$. This function, for instance, can be mean, minimum, maximum, or median. In this paper, $f$ is the minimum function, i.e., a greedy approach. Humans' limited working memory is accounted for by setting a maximum search depth $D_{max}$. If the goal is not found after expanding the graph to depth $D_{max}$, the cost-to-go from a node $n_k$ at depth $D_{max}$ is approximated by $f(CTG'_{k_{list}})$. The model also uses a discount factor $\gamma$ ($0 < \gamma \le 1$). The cost at depth $d$ is weighted by $\gamma^{depth}$. Therefore, the lower the discount factor, the less importance the model gives to the cost at a depth. Discount factor models if a subject is biased towards immediate (local) cost than global cost.   

\subsection{Exploration Metric}

Learning or search tasks in general involves trade-off between exploration (learning new knowledge) and exploitation (using current knowledge to make optimal decisions)~\cite{March1991}. In this paper, the connection matrix extracted from human data is used to quantify exploration behavior. $Q'_{ki}$ gives the number of times the segment associated with the edge $n_k \rightarrow n_i$ is taken by a subject. 
This information is used to determine $M_h$ which represents the number of segments that are taken $h$ times. An exploration metric $EM$ is calculated as follows:
\begin{equation}
EM = \sum_{h=1}^{h=\infty} \left( {M_h \over h} \right) 
\end{equation} 
A large $EM$ corresponds to when a subject explores many different segments only a few times (e.g., once or twice) and a small $EM$ results from a subject taking a subset of edges many times. $EM$ is a measure of exploration behavior of a subject.

\subsection{Extracting Guidance Primitives (GPs)}

\begin{figure}[!htbp]
\centering
\includegraphics[scale=1]{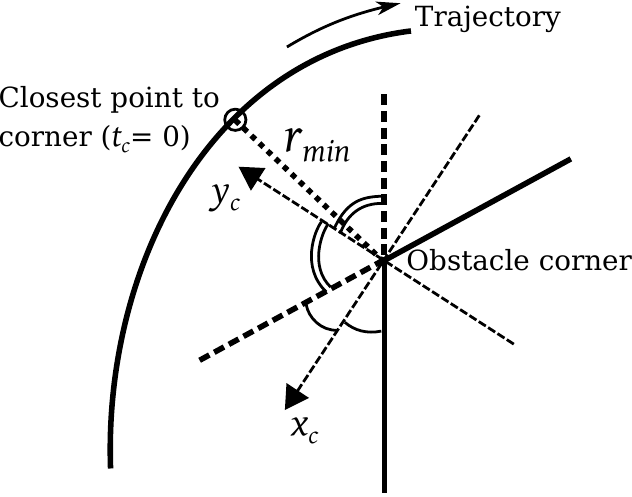} 
\caption{An example trajectory in corner-frame.}
\label{CornerFrame}
\end{figure}

In human data, it is observed that at large distances from obstacle corners subjects mostly travel in straight lines at high speeds. Agent-environment interactions take place when subjects pass close to obstacle corners. As discussed in Section~\ref{sec:introduction}, the hypothesis for task environment learning is that through interactions with the task a pilot learns invariant perceptual and guidance strategies, i.e., guidance primitives~\cite{KongHMSIEEE2013}. 
The analysis of guidance behavior in this paper focuses on trajectory segments in vicinity of corners as agent-environment interactions are high when passing obstacle corners. For this purpose, trajectories are aggregated and described in a common reference frame. Fig.~\ref{CornerFrame} shows the corner-frame used to investigate the guidance primitives. The corner-frame axes are the bisectors of angles formed by walls (boundaries) that meet at the corner. 


First, candidate guidance primitive (GP) segments are extracted as follows. Trajectory segments are transformed into corner frame by translations, rotations, and reflections. Time-origin ($t_c=0$) for a trajectory in corner frame is set at the closest point to the corner (see Fig.~\ref{CornerFrame}). A trajectory segment $s_i$ in corner frame is a sequence of points as follows:
\begin{eqnarray}
s_i = \{..,({x^i_c}_l \ {y^i_c}_l),..\}, \ l \in [1 \ .. \ L], \\ \nonumber
t_c(1)=-T, \ t_c(L)=T,
\end{eqnarray}
where $2T$ is the time-duration of trajectory segment considered for subsequent analysis of candidate GPs. $L$ is the number of discrete points in time-duration $2T$.  
Distance $d^{ij}_{s}$ between two trajectories $s_i$ and $s_j$ is defined as follows:
\begin{eqnarray}
\label{dij_pi}
d^{ij}_{s} = \sum_{l=1}^{l=L} w \sqrt{({x^i_c}_l - {x^j_c}_l)^2 + ({y^i_c}_l - {y^j_c}_l)^2}, \\ \nonumber
w = 1 - {|t_c(l) - T| \over 2T}.
\end{eqnarray}
The distance in Eq.~\ref{dij_pi} is based on points that have the same time-instant, which distinguishes trajectories that are similar in geographical space but have different motion behavior (e.g., speed and turnrate). Points on trajectory segments are weighed based on how far they are from closest point to corner. 

Distance $d^{IJ}_c$ between two clusters $C^{I}_{s}$ and $C^{J}_{s}$ is the average distance between all pairs of trajectories $s_i \in C^{I}_{s}$ and $s_j \in C^{J}_{s}$ as follows:
\begin{equation}
d^{IJ}_c = {1 \over {|C^{I}_{s}| |C^{J}_{s}|}} \sum_{i=1}^{i=|C^{I}_{s}|} \sum_{j=1}^{j=|C^{J}_{s}|} d^{ij}_{s},
\label{dIJ_c}
\end{equation}
where $|C^{I}_{s}|$ is the number of trajectories in $I^{th}$ cluster, i.e., $C^{I}_{s}$. 
Trajectories are clustered using the bottom up hierarchical clustering. 
Each trajectory starts as a single cluster. As moving up the hierarchy, two closest (minimum $d^{IJ}_c$) clusters are merged. The process is repeated until a specified number of clusters is achieved.


\section{Results and Analysis}
\label{sec:results}
This section presents an analysis of human data using the framework proposed in the previous section. First, it presents general observations that focus on planning, exploration, convergence in CTG at subgoals, and evolution in control and gaze behavior with environment learning. Finally, the section presents a quantitative analysis of guidance primitives associated with interaction patterns that emerge with environment learning.

\subsection{Planning (Decision-Making)}

\begin{figure}[htbp]
\centering
\includegraphics[scale=0.6]{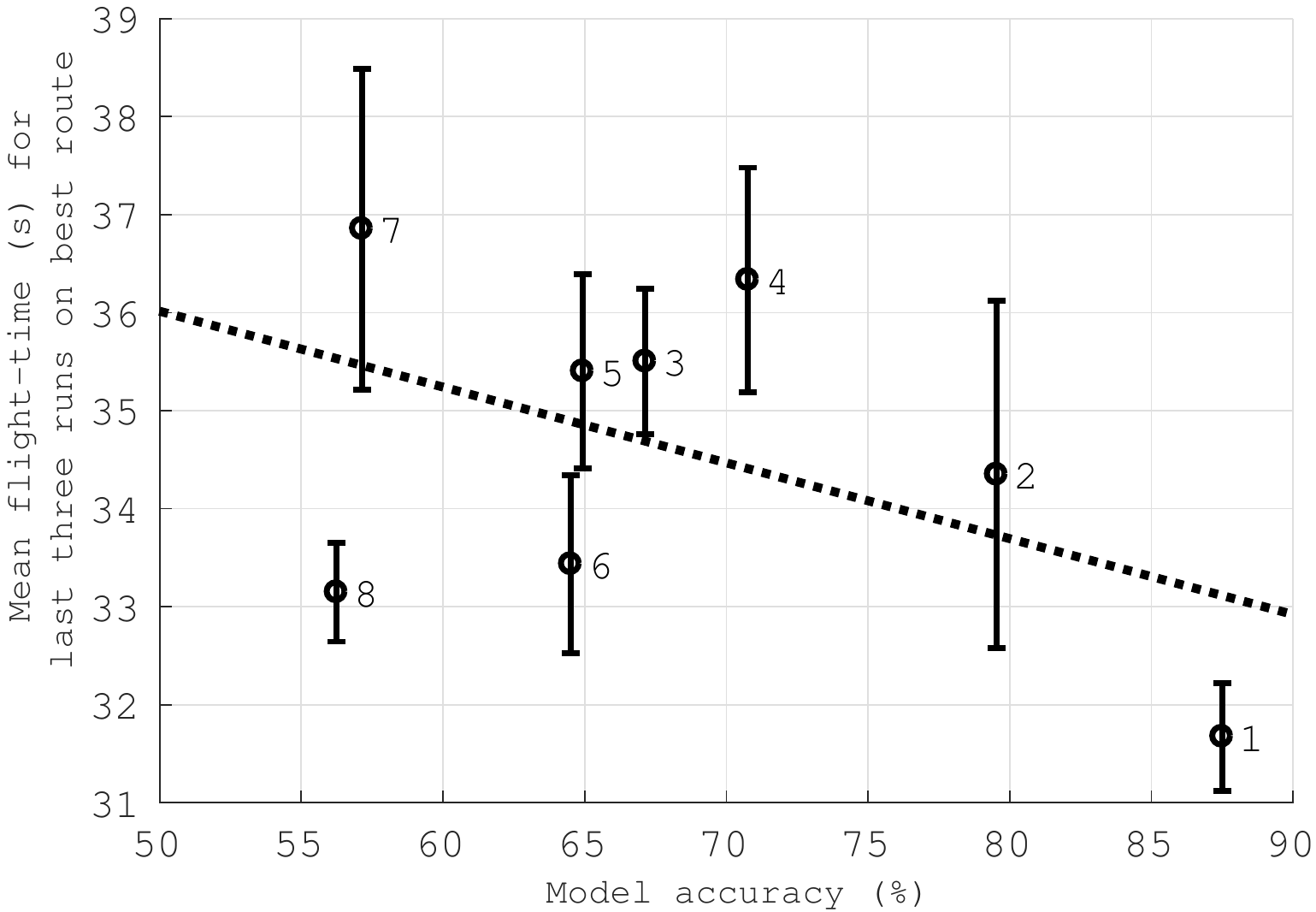} 
\caption{Flight-time on a subject's best route vs the model accuracy.}\label{FT_Accuracy}
\end{figure}

Figure~\ref{FT_Accuracy} shows the decision model accuracy (for $D_{max} = \infty$ and $\gamma = 1$) and mean and standard deviation of flight-time for each subject's last three runs on their best route. Model accuracy and flight-time correspond to operator rationality and performance, respectively. It is reasonable to assume that a better model accuracy should result in a lower flight-time. The best line fit between model accuracy and flight-time is shown by the dotted line in Fig.~\ref{FT_Accuracy}. Subject \# 1 is the best, i.e., maximum accuracy (87.5 $\%$) and best flight-time (mean and standard deviation are 31.7 s and 0.5 s, respectively). 
Subject \# 8 is an outlier and achieves the second best flight-time (mean and standard deviation are 33.1 s and 0.5 s, respectively) despite the worst model accuracy (56.3 $\%$). 
Subject \# 7 shows the worst flight-time (mean and standard deviation are 36.9 s and 1.6 s, respectively) and second worst model accuracy (57.1 $\%$). 

\begin{figure}[htbp]
\centering
\subfigure[]{\includegraphics[width = 2in]{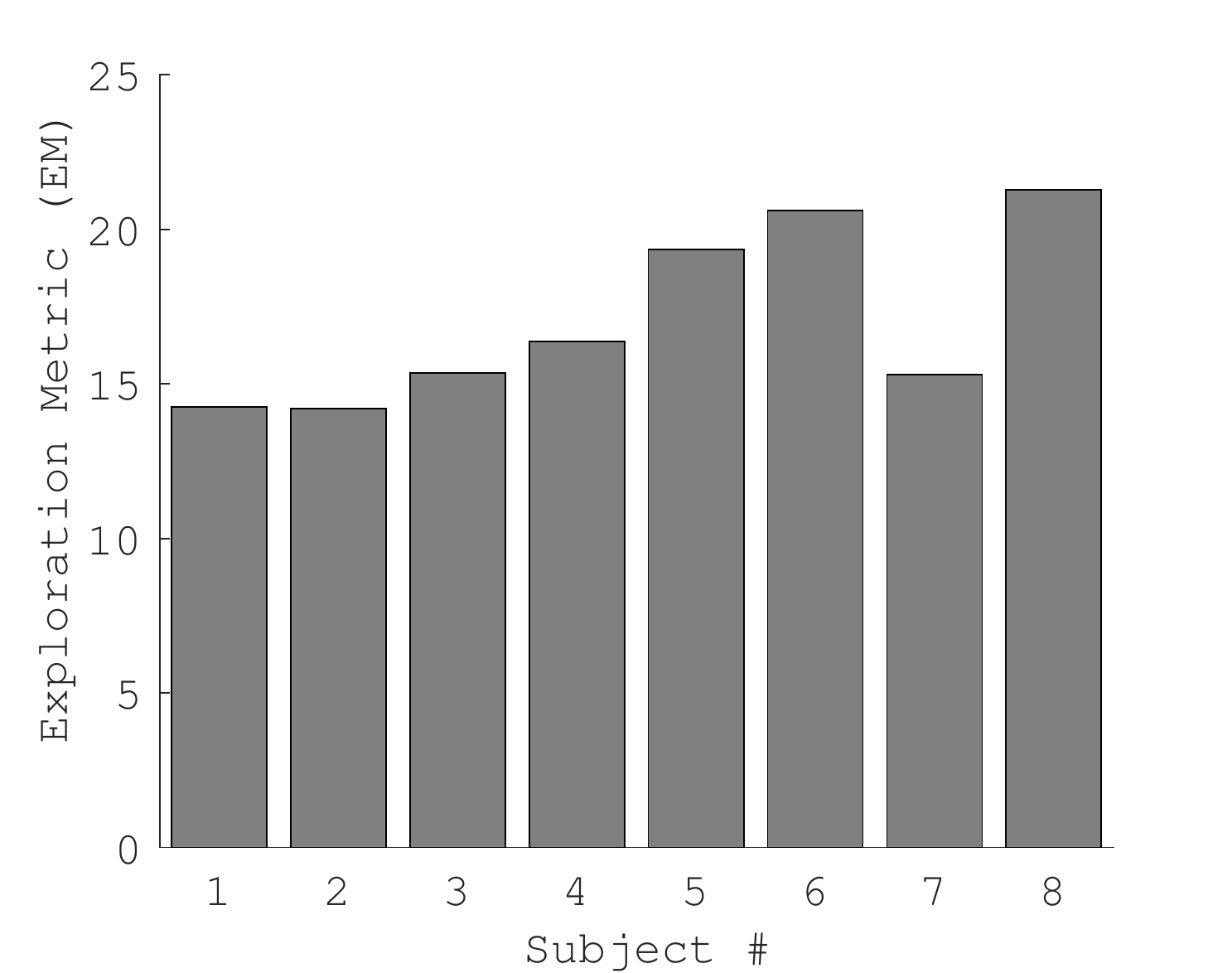}}
\subfigure[]{\includegraphics[width = 4in]{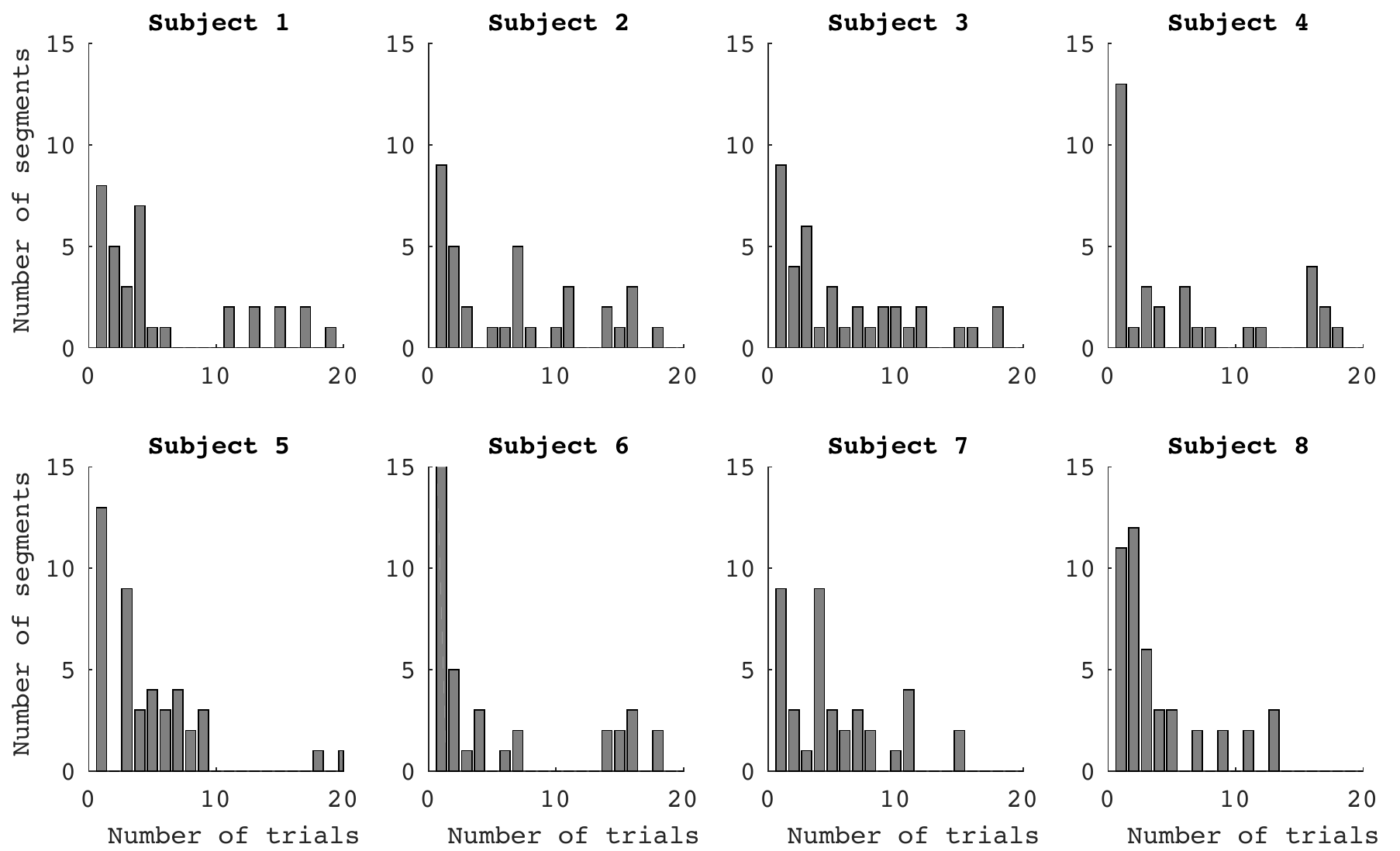}}
\caption{(a) Exploration metric ($EM$) and (b) Distribution of segments based on trial frequency.}
\label{EM}
\end{figure}

\subsubsection{Exploration vs. Exploitation}

Figure~\ref{EM}(a) shows the exploration metric ($EM$) for all subjects. Subject \# 8 has the largest $EM=21.3$. Figure~\ref{EM}(b) shows the distribution of segments based on their trial frequency. Subject \# 8 tries several segments few times unlike other subjects. This high exploration tendency of subject \# 8 may be a reason why the subject has the lowest model accuracy (56.3 $\%$) despite the second best flight-time (mean is 33.1 s) on its best route. 

\subsubsection{Visibility}

\begin{figure}[!htbp]
\centering
\includegraphics[scale=0.6]{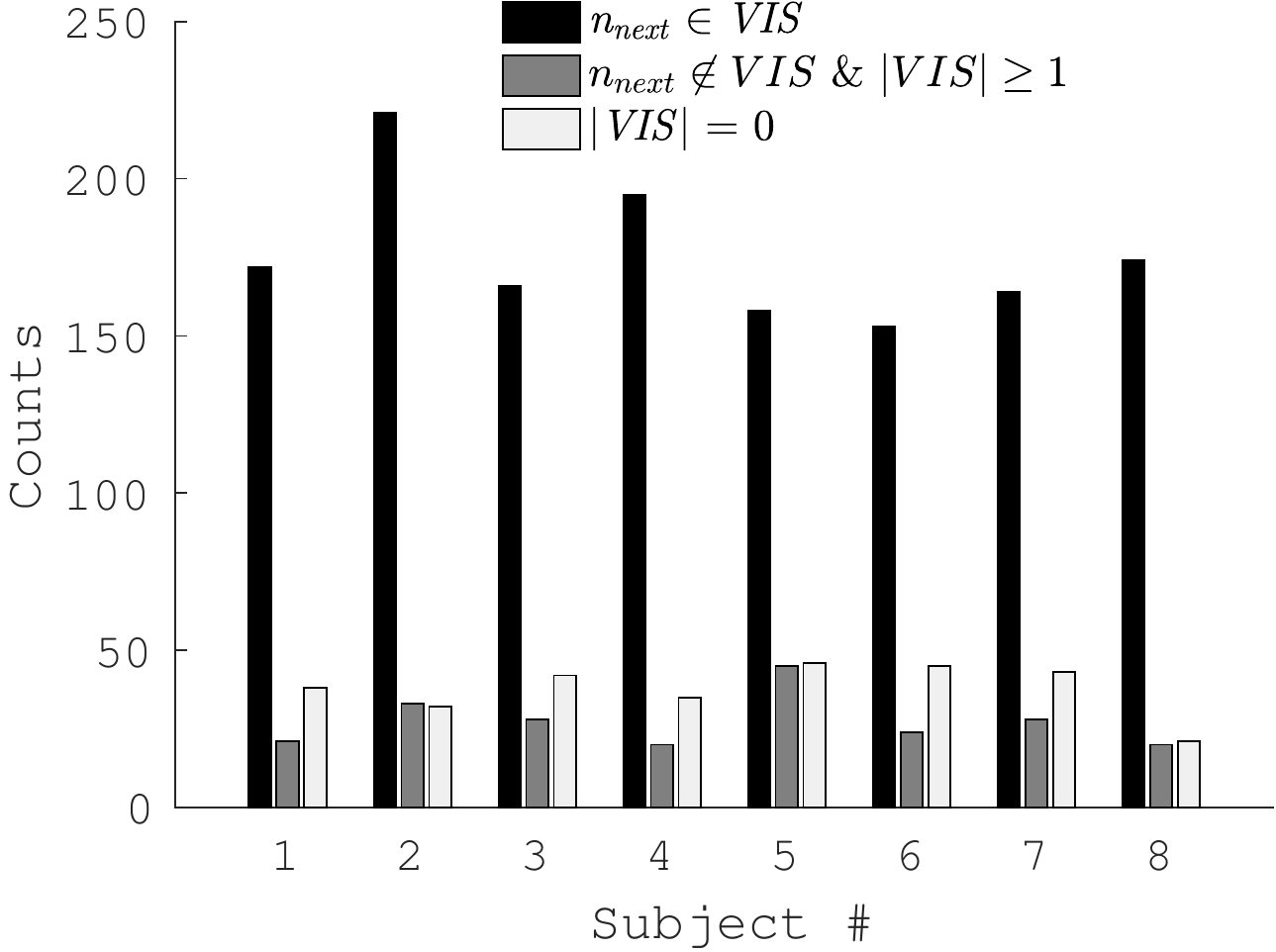} 
\caption{Number of occurrences for $n_{next} \in VIS$, $n_{next} \not\in VIS$, and no visible nodes for all subjects.}
\label{VIS}
\end{figure}

The simulation system models the environment that is within the field of view (60$\degree$) of an operating subject. A node is visible if it is in the field of view and not obscured or hidden by obstacles. Figure~\ref{VIS} shows the number of occurrences that the next node $n_{next}$ chosen by a subject is $\in VIS$, $\not\in VIS$, or $VIS=\{ \}$. It can be seen that subjects often (mean frequency is 93 \% for all subjects) choose visible nodes when there is any. 

\subsection{Environment Learning}

\begin{figure}[htbp]
\centering
\subfigure[]{\includegraphics[width = 3.2in]{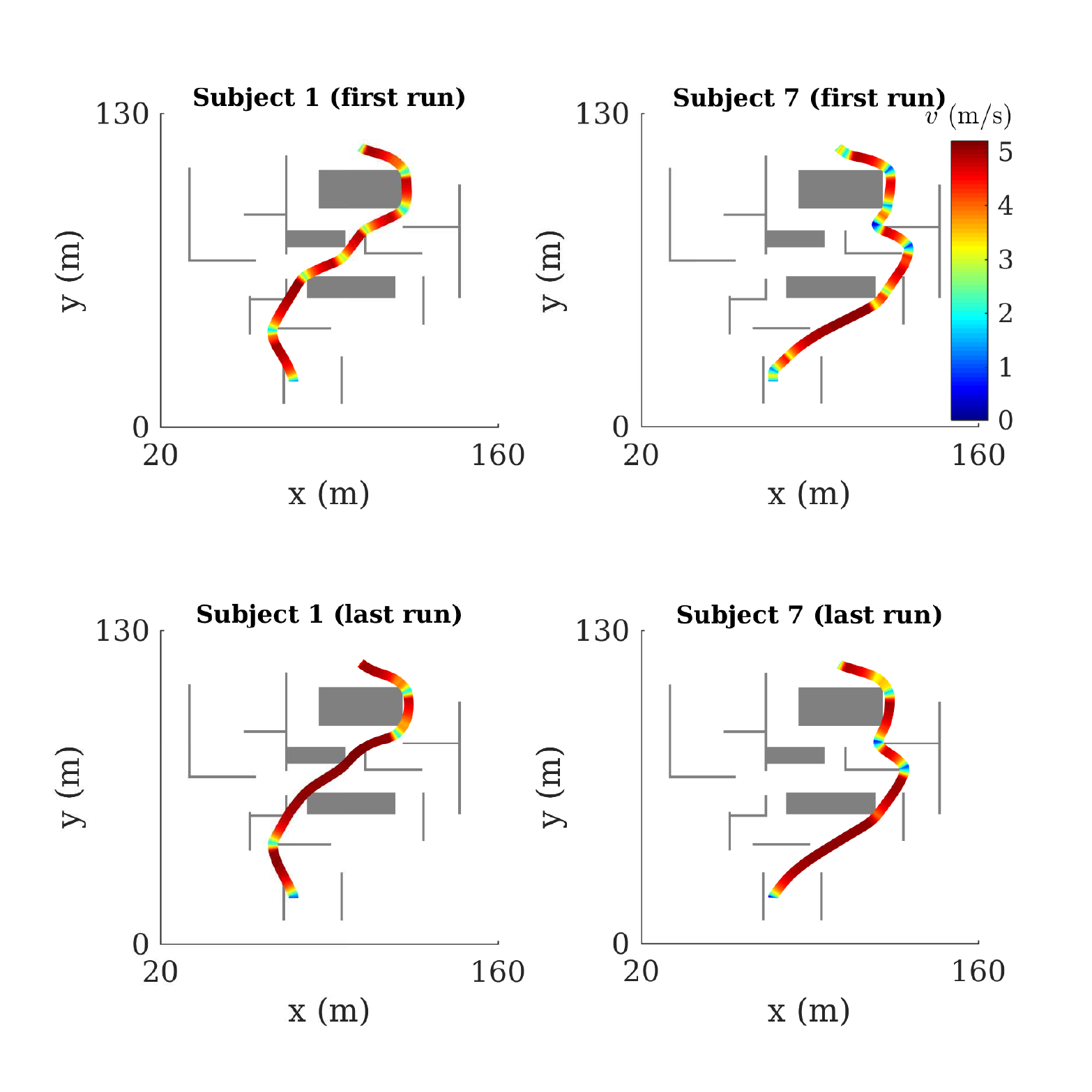}}
\subfigure[]{\includegraphics[width = 3.2in]{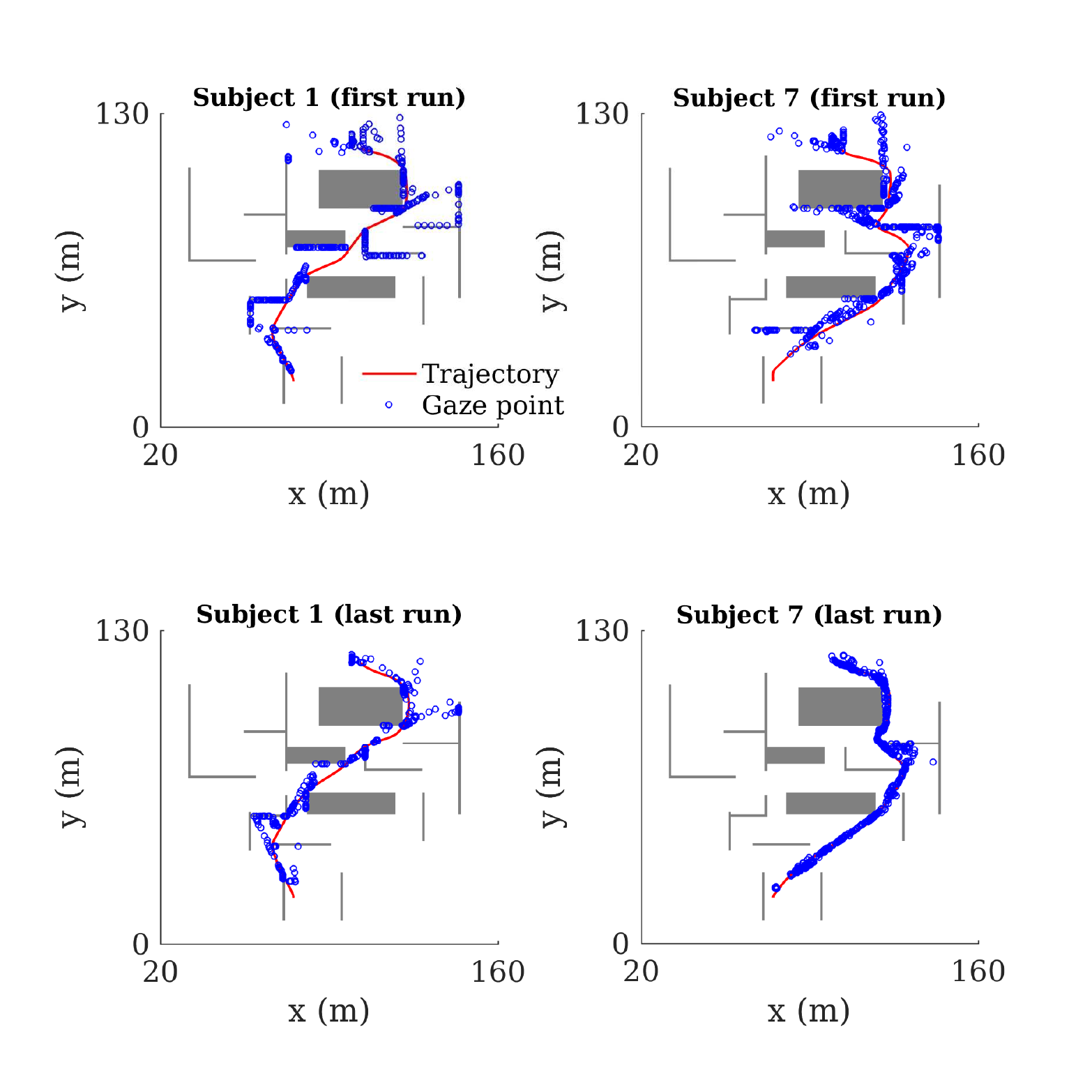}}
\caption{(a) Speed and (b) Gaze trajectories for first and last runs on best routes of subjects \# 1 and \# 7.}
\label{Speed_Gaze}
\end{figure}

This section compares subjects \# 1 and \# 7 who give best and worst flight-times, respectively, for environment learning analysis. Figure~\ref{Speed_Gaze}(a) shows speed time-histories for first and last runs on best routes of subjects \# 1 and \# 7. In starting runs, subjects slow down as they approach any obstacle corner (or subgoal $g_k$) because parent subgoal $(g_k)_p$ and therefore subgoal velocity $[v^{g_k} \ \psi^{g_k}]$ are unknowns in starting runs. As the environment is learned, subgoal network and velocities are learned. In later runs, subjects reduce speed, when approaching a subgoal $g_k$, based on turning required to align with the next (parent) subgoal $(g_k)_p$.  

\begin{figure}[htbp]
\centering
\subfigure[]{\includegraphics[width = 3.2in]{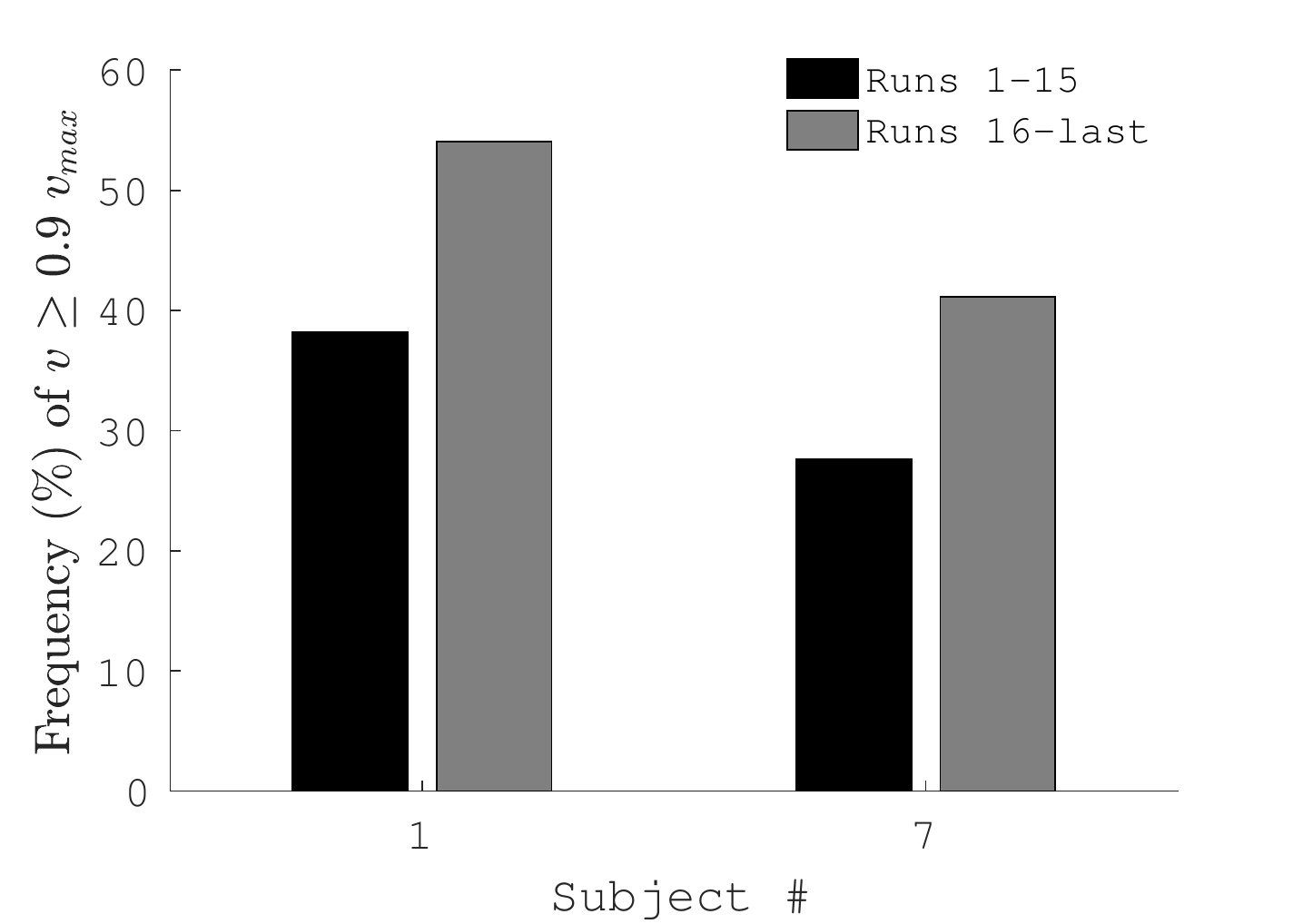}}
\subfigure[]{\includegraphics[width = 3.2in]{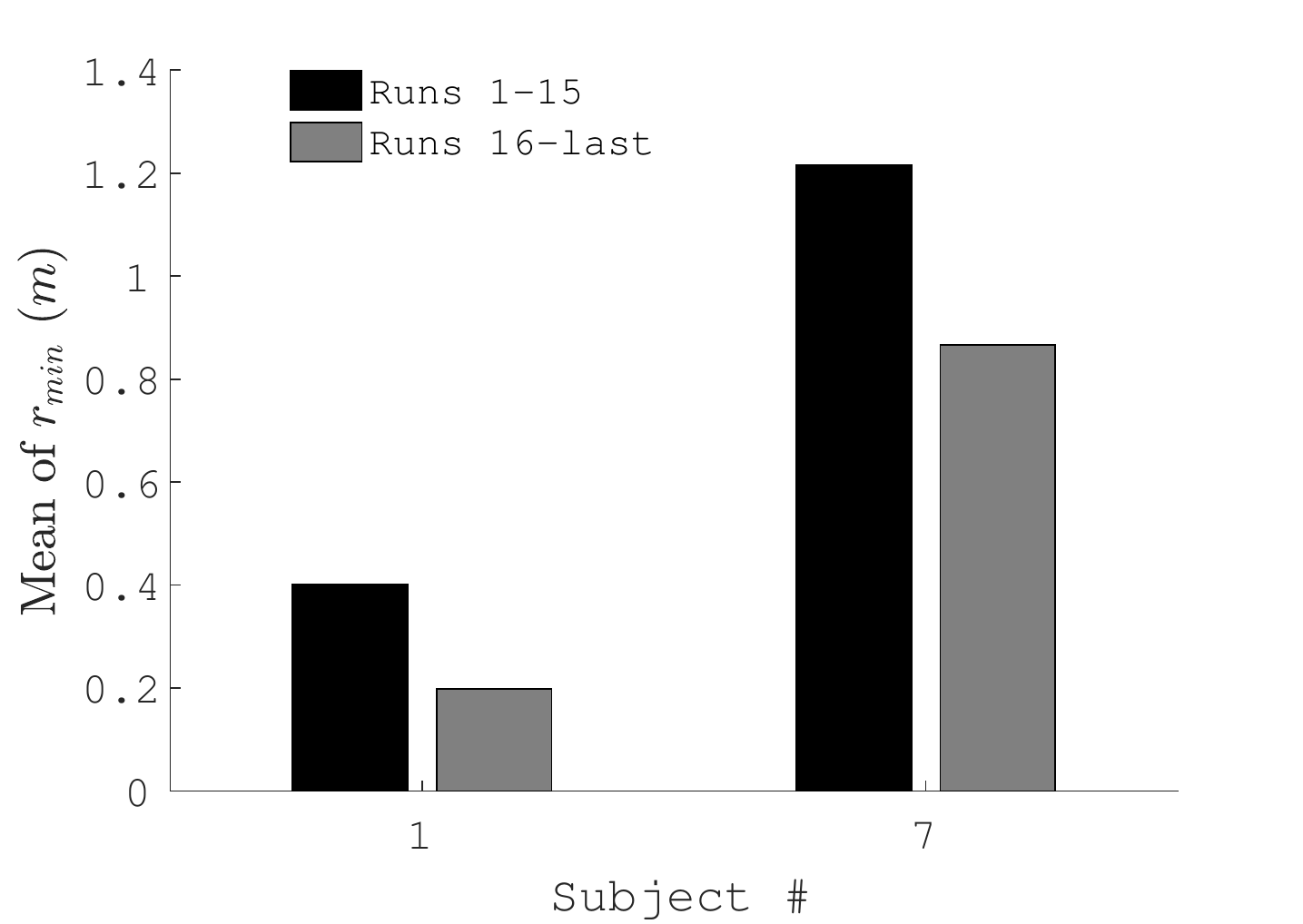}}
\caption{(a) Frequency of high-speeds near corners and (b) Mean $r_{min}$ for starting (1-15) and final (16-last) runs for subjects \# 1 and \# 7.}
\label{HighSpeed_Rmin}
\end{figure}

Figure~\ref{HighSpeed_Rmin}(a) shows frequencies of high-speeds ($\ge$ 90 $\%$ of $v_{max}$) for starting (1-15) and final (16-last) runs for subjects \# 1 and \# 7. The frequencies are computed using trajectory data near corners (within time-window $T$ = 2$\tau$ from a corner, where $\tau=1.13$ $s$ is the time-constant for the vehicle command-to-speed model). For subject \# 1, the frequency of high-speeds increases from 38.2 $\%$ to 54.0 $\%$ from starting to final runs. For subject \# 7, the frequency increases from 27.7 $\%$ to 41.1 $\%$. Figure~\ref{HighSpeed_Rmin}(b) shows the mean minimum distance ($r_{min}$) from obstacle corners for starting and final runs for the both subjects. In final runs, mean $r_{min}$ for subjects \# 1 and \# 7 are 0.2 $m$ and 0.9 $m$, respectively. These results suggest that Subject \# 7 shows higher obstacle avoidance behavior than subject \# 1. 

\subsubsection{Gaze}

\begin{figure}[htbp]
\centering
\includegraphics[scale=0.6]{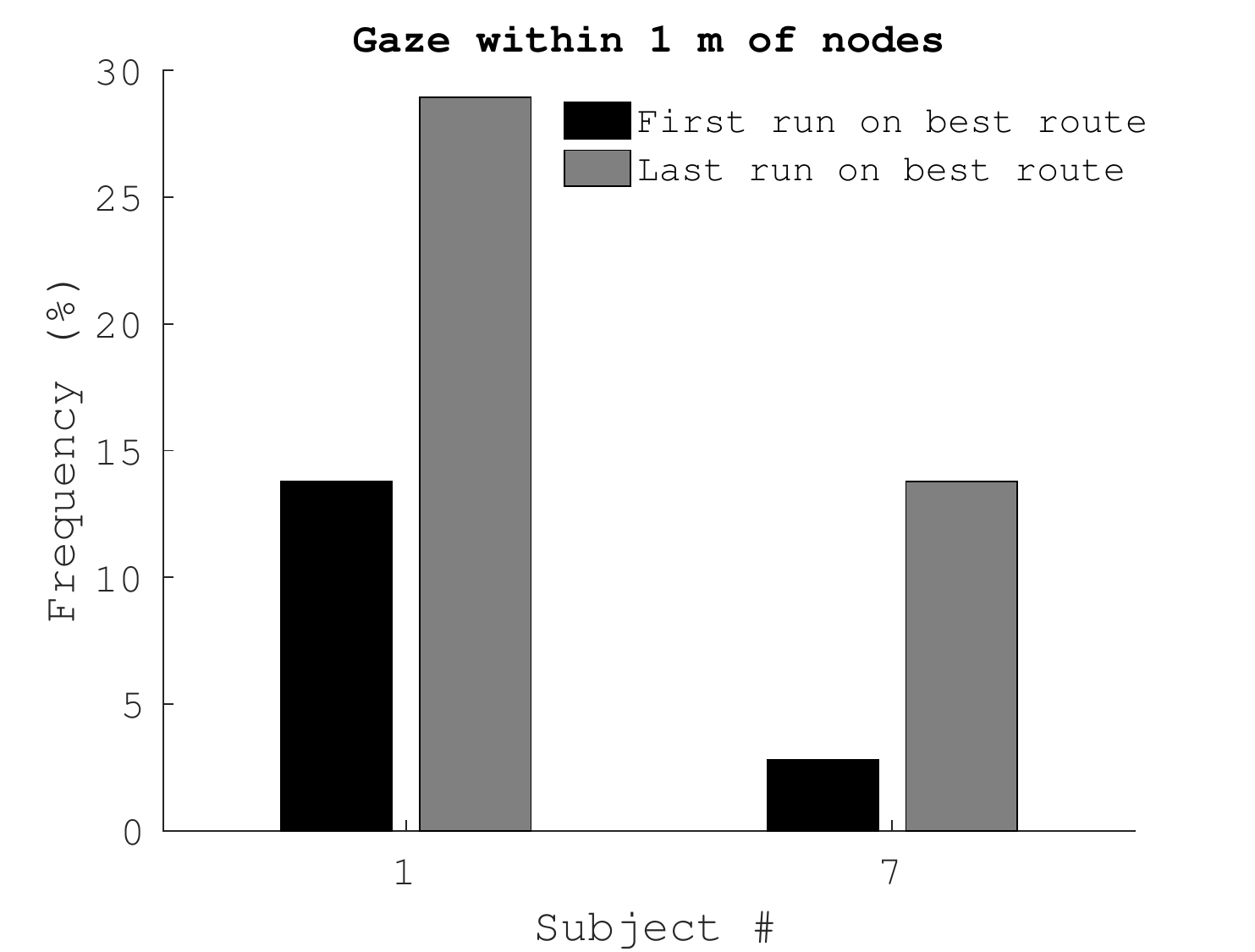} 
\caption{Frequency of gaze within 1 $m$ of corners in first and last runs on best routes of subjects \# 1 and \# 7.}
\label{GazeAtCorners}
\end{figure}

Figure~\ref{Speed_Gaze}(b) shows gaze trajectories for first and last runs on best routes of subjects \# 1 and \# 7. Figure~\ref{GazeAtCorners} shows the frequency of gaze within 1 $m$ of obstacle corners, i.e., subgoal heuristics, for the runs shown in Fig.~\ref{Speed_Gaze}. 
Visual attention in starting runs is scattered (e.g., regularly scanning sideways) for both subjects. In the last run, subject \# 1 primarily (28.9 $\%$ of total time) focuses gaze near obstacle corners. Subject \# 7 attends to obstacle corners with almost half the frequency (13.8 $\%$ in the last run) of subject \# 1, and he/she focuses gaze at future points on the path. An explanation for such gaze behavior of subject \# 7 is that the subject is occupied with stabilizing the vehicle on a reference path due to his/her novice control skills, which is showed later in the analysis of guidance primitives. 

\subsubsection{CTG at Subgoals}

\begin{figure}[htbp]
\centering
\includegraphics[scale=0.6]{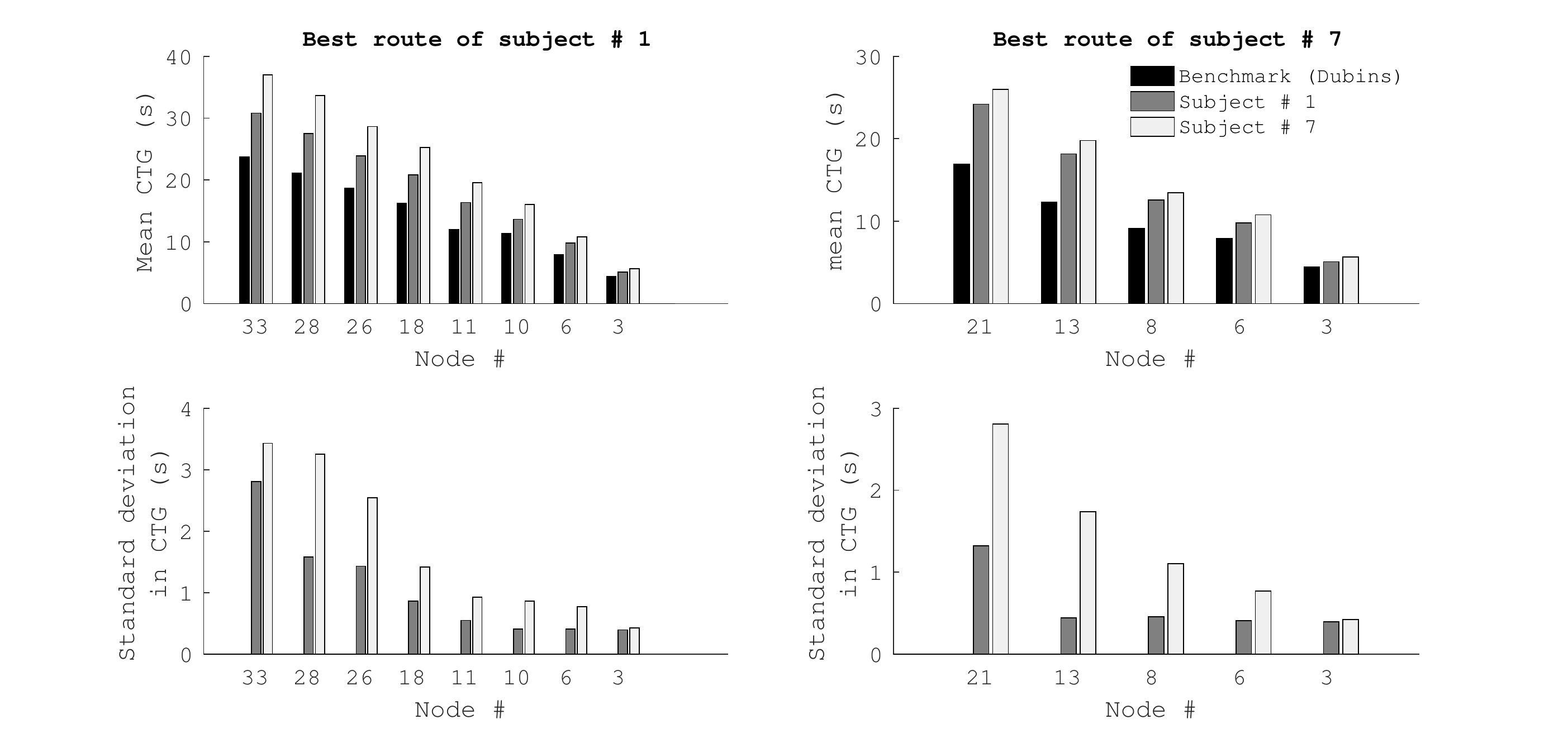} 
\caption{Benchmark, mean, and standard deviation of CTG for subjects \# 1 and \# 7 at nodes on their best routes.}
\label{CTGNodes}
\end{figure}

Figure~\ref{CTGNodes} shows the benchmark CTG and mean and standard deviation of CTG for subjects \# 1 and \# 7 at nodes on their best routes. The average gap between the benchmark CTG and subject \# 1's mean CTG is 26.5 $\%$. For subject \# 7, the gap is 49.3 $\%$. Mean standard deviation in CTG's at nodes for subjects \# 1 and \# 7 are 5.4 and 7.3 $\%$, respectively. Subject \# 1 shows better convergance in CTG at subgoals (nodes) than subject \# 7.

\subsection{Guidance Primitives (Quantitative Analysis)}

\begin{figure*}[!htbp]
\centering
\includegraphics[scale=0.54]{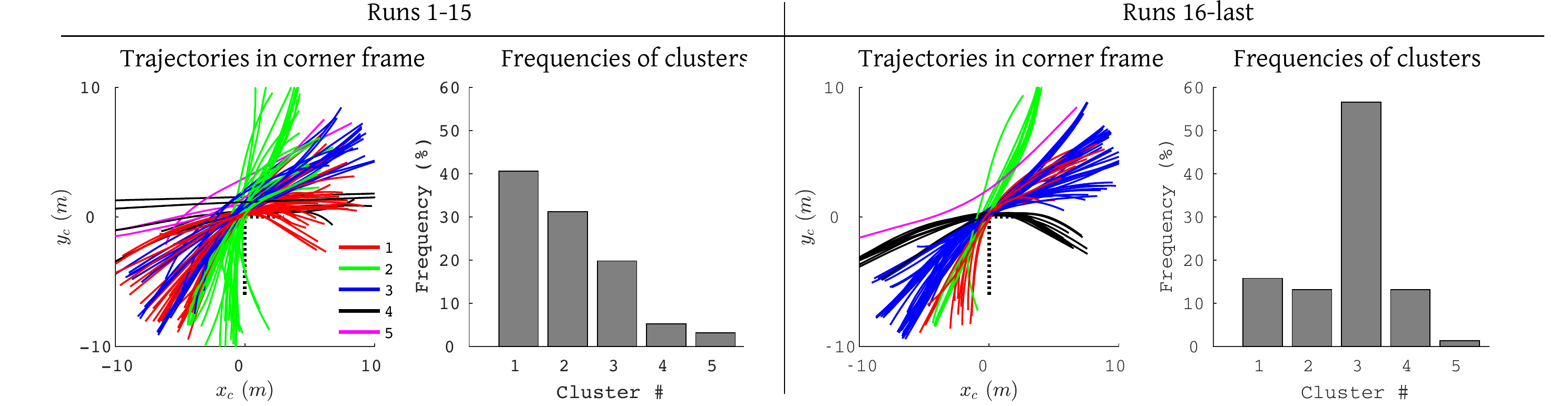} 
\caption{Subject \# 1: trajectories in corner frame and clusters' frequencies for runs 1-15 and 16-last.}
\label{S1_trajs_clusters}
\end{figure*}

\begin{figure*}[!htbp]
\centering
\includegraphics[scale=0.54]{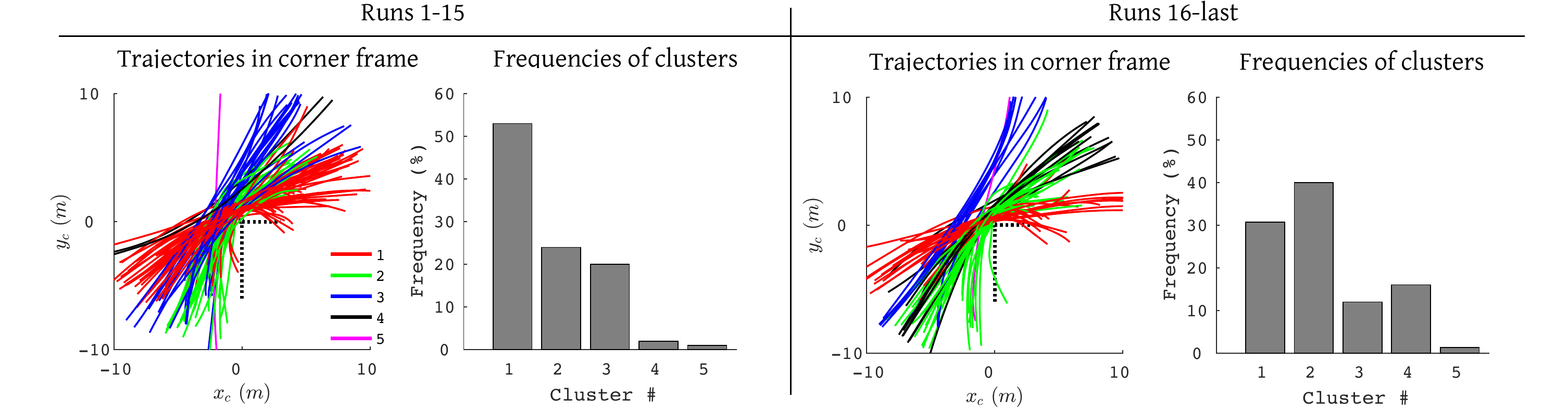} 
\caption{Subject \# 7: trajectories in corner frame and clusters' frequencies for runs 1-15 and 16-last.}
\label{S7_trajs_clusters}
\end{figure*}

\begin{figure*}[!htbp]
\centering
\includegraphics[scale=0.6]{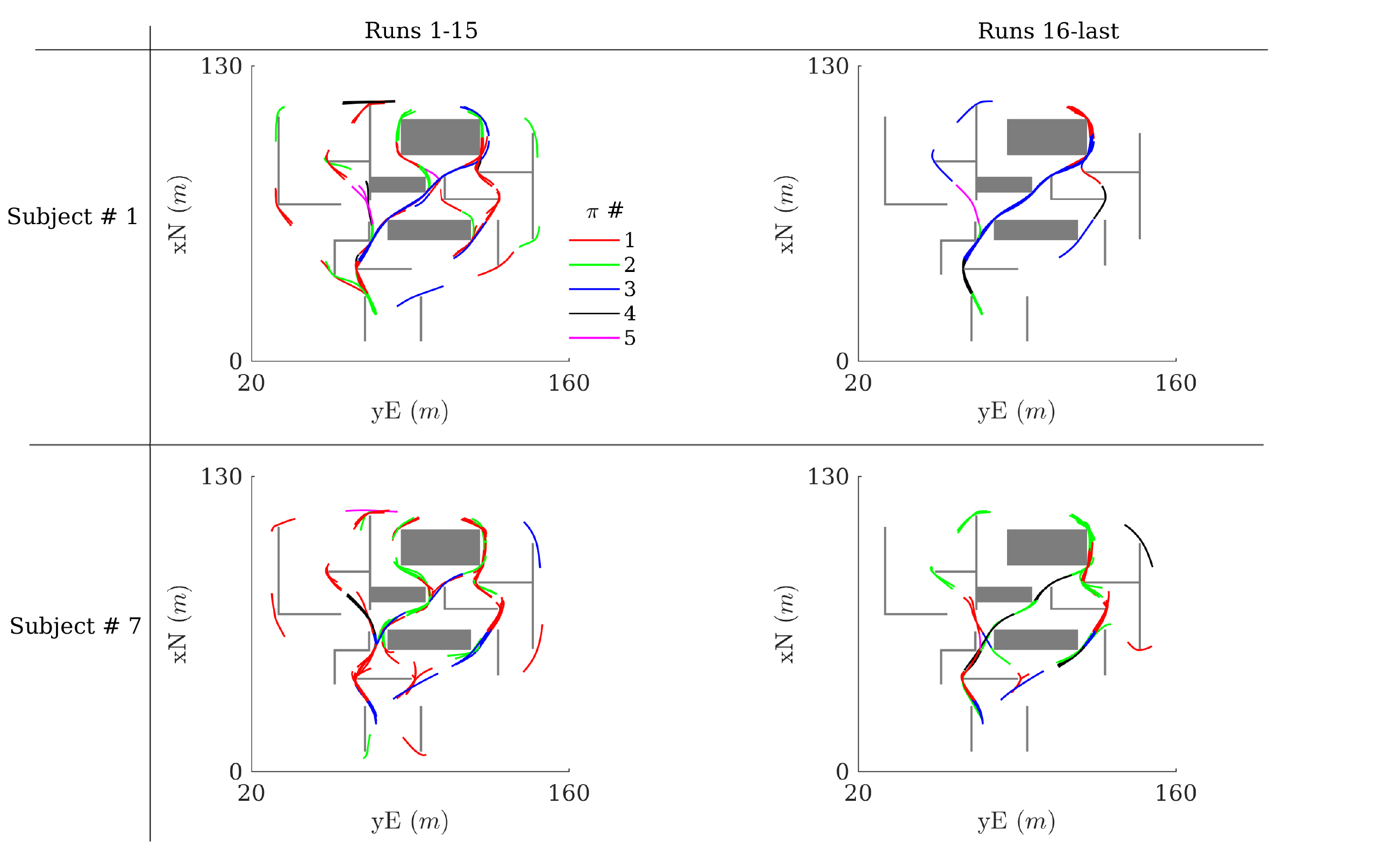} 
\caption{Subjects A and B: trajectory clusters 1-5 in global environment for runs 1-15 and 16-last.}
\label{S17_global_GP}
\end{figure*}

Figures~\ref{S1_trajs_clusters} and \ref{S7_trajs_clusters} show trajectory segments in the corner frame for runs 1-15 and 16-last for subjects \# 1 and \# 7, respectively. Time-window $T$ is 2$\tau$ where $\tau=1.13$ $s$ is the time-constant for the vehicle command-to-speed model. The trajectories are divided into five clusters ($\pi_i, i \in [1 \ 5]$) using hierarchical clustering (Eqs.~\ref{dij_pi} and \ref{dIJ_c}). In runs 1-15, clusters are numbered in decreasing order of frequencies. In runs 16-last, clusters are numbered according to their similarity with the clusters in runs 1-15. The similarity between two clusters is measured  as the average distance between all pairs of trajectory segments in the clusters (Eq.~\ref{dIJ_c}). Figure~\ref{S17_global_GP} shows the trajectory segments in the global environment for both the subjects.

\begin{figure*}[!htbp]
\centering
\includegraphics[scale=0.43]{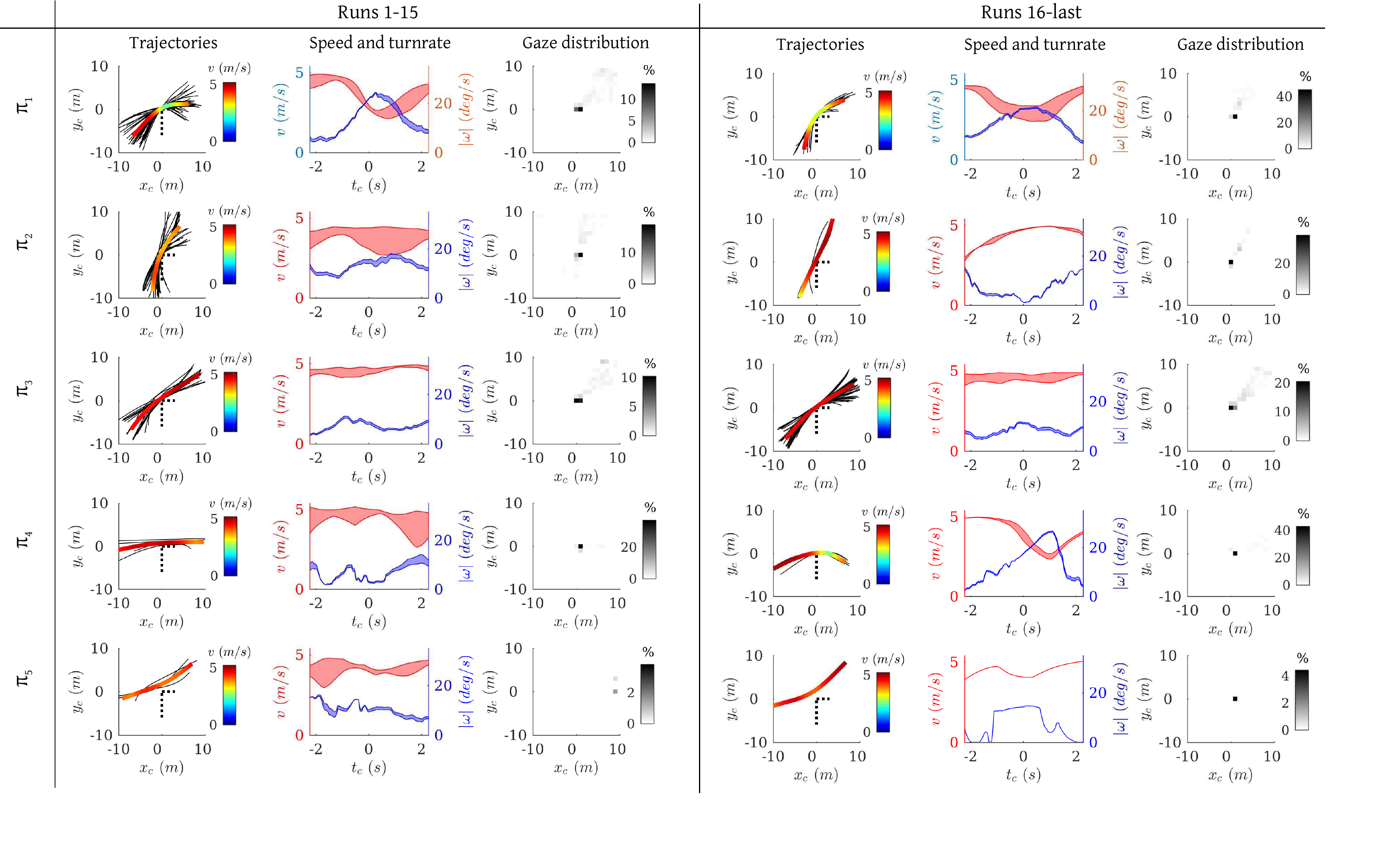} 
\caption{Subject \# 1: trajectories, speed, turnrate, and gaze distribution for clusters 1 to 5 for runs 1-15 and 16-last.}
\label{S1_clusters_speed_tr_gaze}
\end{figure*}

\begin{figure*}[!htbp]
\centering
\includegraphics[scale=0.43]{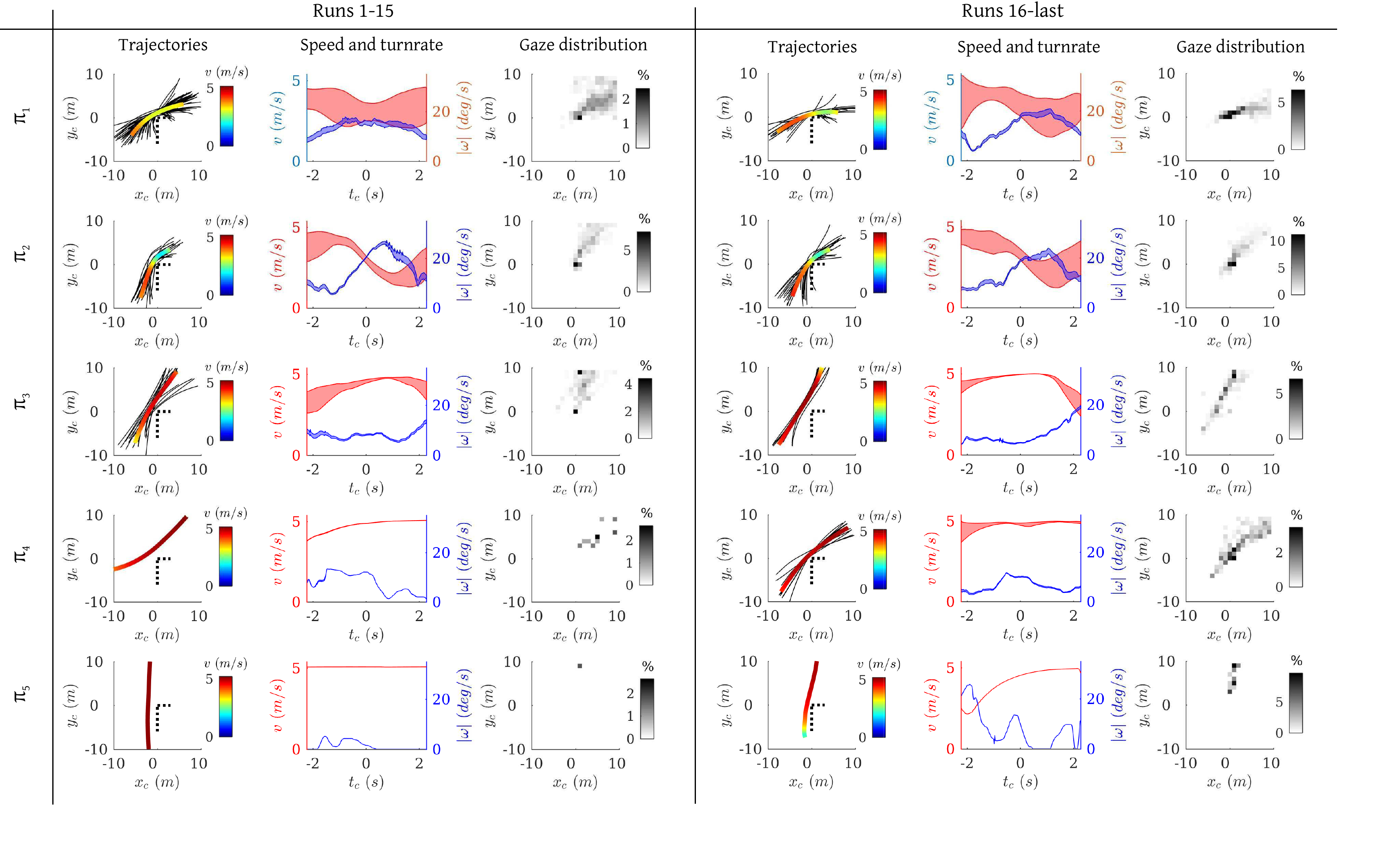} 
\caption{Subject \# 7: trajectories, speed, turnrate, and gaze distribution for clusters 1 to 5 for runs 1-15 and 16-last.}
\label{S7_clusters_speed_tr_gaze}
\end{figure*}

The frequencies of clusters in runs 1-15 and 16-last for subjects \# 1 and \# 7 are shown in Figs.~\ref{S1_trajs_clusters} and \ref{S7_trajs_clusters}, respectively. 
For subject \# 1, clusters are not distinct in runs 1-15. The behavior follows more distinct clusters in runs 16-last (see trajectories in Fig.~\ref{S1_trajs_clusters}). In runs 1-15, there are three dominant clusters with frequencies of 40.6, 31.3, and 19.8 $\%$, respectively. In runs 16-last, there is one dominant mode with the frequency of 56.6 $\%$. For subject \# 7, trajectories in runs 16-last are spread across clusters. Subject \# 1 has a guidance primitive library ($\Pi$) with better differentiated behaviors than subject \# 7. 

Figures~\ref{S1_clusters_speed_tr_gaze} and \ref{S7_clusters_speed_tr_gaze} show trajectories, mean trajectory (colored based on mean speed value), time-histories of mean speed and turnrate, and gaze distribution for the clusters for runs 1-15 and 16-last for subjects \# 1 and \# 7, respectively. For a cluster, overall mean speed $V$ and uncertainty in speed profile $U_v$ are computed as follows:
\begin{eqnarray}
V = {{\int_{-T}^{T} w v_{m} dt_c} \over {{\int_{-T}^{T} w dt_c}}}, \\ \nonumber
U_v = {{\int_{-T}^{T} w {\sigma}_v  dt_c} \over {{\int_{-T}^{T} w dt_c}}}, \\ \nonumber
w = 1 - {|t_c - T| \over 2T},
\end{eqnarray} 
where $v_{m}$ and $\sigma_v$ are mean and standard deviation in speed, respectively. $V$ and $U_v$ for subjects \# 1 and \# 7 for the clusters (guidance primitives: $\pi_i, i \in [1 \ 5]$) for runs 1-15 and 16-last are shown in table~\ref{Table2}. The table also shows the $V$ and $U_v$ for the guidance primitive library $\Pi$, which are weighted sum of $V$ and $U_v$ for clusters $\pi_i$'s based on their frequencies, in runs 1-15 and 16-last for the both subjects. The mean speed for subject 1 in runs 16-last is 4.3 $m/s$ with the standard deviation of 0.2 $m/s$, which are 3.7 $m/s$ and 0.5 $m/s$, respectively, for subject \# 7. 

Subject \# 1 shows consistent (repeatable) control behavior unlike subject \# 7. This observation supports that subject \# 1 has consolidated the behavior in his/her memory. Also, the behavior consolidated in subject \# 1's memory is effective and safe, which are supported by high speeds used by the subject (Figs.~\ref{HighSpeed_Rmin}(a) and Fig.~\ref{S1_clusters_speed_tr_gaze}) and close distances to corners (Fig.~\ref{HighSpeed_Rmin}(b)), respectively.


Gaze distribution in Figs.~\ref{S1_clusters_speed_tr_gaze} and \ref{S7_clusters_speed_tr_gaze} are computed using gaze data from $t_c=-T$ to $t_c=0$ because corner is not visible beyond $t_c=0$. In runs 1-15,
subject \# 1 focuses gaze near corners with the frequency of 10-20 $\%$. In runs 16-last, the frequency increases to 20-40 $\%$, which is almost four times the frequency (5-10 $\%$) of subject \# 7. Subject \# 7 looks at future points on the path instead of focusing at corners, which is consistent with observations in Fig.~\ref{Speed_Gaze}(b).

Subject \# 1 who achieves lower flight-time and better differentiated and converged guidance primitives than subject \# 7, focuses gaze at corners. There are two possible reasons for subject \# 1's focus at corners. One reason is based on bottom-up visual processing, i.e., corners are salient visual features. Another reason is top-down planning strategy where corners are heuristics for subgoals. Subject \# 1's gaze focus at corners in $\pi_1$ is 40-45 $\%$ whereas it is almost the half (20 $\%$) in $\pi_3$. The trajectories in $\pi_1$ involve higher turning of vehicle around the corner than the trajectories in $\pi_3$. This observation supports that the attention at corners is not only due to saliency but also because corners serve as subgoal heuristics.

\begin{table*}[!htbp]
\caption{Overall mean ($V$) and uncertainty ($U_v$) of speed profile for clusters \# 1 to \# 5 (guidance primitives: $\pi_i, i \in [1 \ 5]$) and all clusters together (guidance primitive library $\Pi$) for subjects \# 1 and \# 7 for runs 1-15 and 16-last.}
\begin{center}
\begin{tabular}{ |c|c|c|c|c|c|c|c|c| } 
\hline
Runs & & $\pi_1$ & $\pi_2$ & $\pi_3$ & $\pi_4$ & $\pi_5$ & $\Pi$ \\ \hline \hline
1-15 & Subject \# 1: $V$ ($U_v$) $m/s$ & 3.4(0.3) & 3.8(0.6) & 4.6(0.2) & 4.6(0.3) & 4.1(0.3) & 3.8(0.4) \\ 
16-last & Subject \# 1: $V$ ($U_v$) $m/s$ & 3.3(0.4) & 4.6(0.0) & 4.6(0.2) & 3.8(0.2) & 4.4(0.0) & 4.3(0.2) \\ 
1-15 & Subject \# 7: $V$ ($U_v$) $m/s$ & 3.2(0.8) & 3.0(0.3) & 4.4(0.2) & 4.8(0.0) & 5.1(0.0) & 3.4(0.5) \\ 
16-last & Subject \# 7: $V$ ($U_v$) $m/s$ & 3.4(0.8) & 3.2(0.6) & 4.8(0.1) & 4.8(0.1) & 4.3(0.0) & 3.7(0.5)\\ 
\hline
\end{tabular}
\label{Table2}
\end{center}
\end{table*}





\section{Conclusions}
\label{sec:conclusions}


\subsection{Contribution}

This paper extended the prior concept of interaction patterns to formulate hypothesis about environment learning in guidance tasks in unknown obstacle fields. The paper presented a graph learning model based on patterns in sensory-motor behavior in interaction with the spatial environment and task elements, to analyze human environment learning. 
The subgoal graph knowledge of a subject is assessed by trajectories over successive trials. An optimal graph search method is applied to evaluate human planning of subgoals. The model allows testing an operator's rationality and accuracy of the model.

The graph representation of task environment enables formal assessment of three aspects of human task learning: 1) task environment structure (subgoal graph), 2) task performance (cost-to-go across graph), and 3) range of guidance primitives. 
The task environment structure is represented by  connectivity among subgoals. The task performance and improvement with environment learning are assessed by tracking the convergence of cost-to-go at subgoals over successive trials. The guidance primitives that emerge as a result of task environment learning over multiple trials are extracted using a clustering method. 

\subsection{Specific Insights about Human Spatial Behavior}

Proficient subjects demonstrate highly repeatable control behavior over vehicle dynamics and its interaction with the spatial environment. These subjects exhibit clearly formed interaction patterns. The interaction patterns allow subjects to focus their attention on the high-level elements of the task such as subgoals needed to elaborate plans and process relevant environment elements. In contrast, unskilled subjects are mostly focused on basic vehicle controls. Therefore they allocate most of their attention to the low-level functions such as stabilizing the vehicle along a path and avoiding collision. 

The interaction patterns aid planning and ultimately learning, because the largely automated performance of guidance behavior enable filtering the information that is relevant to the execution but is not relevant to the larger task specification, and extract information elements that are relevant to learning the task at hand. This suggest that the interaction patterns are assimilated in procedural memory similar to other sensory-motor patterns studied in human and animal motor control.

\subsection{Application}

The subgoal graph framework, decision-making model, and the gained knowledge is currently being applied to the design of an autonomous guidance algorithm. The algorithm uses a sparse representation, i.e., subgoal graph, of task environment. The graph representation is learned over successive trials. The statistics of cost-to-go at subgoals is used to decide exploration vs. exploitation. A goal is to simulate the autonomous guidance system to verify the emergence of guidance primitives with environment learning as observed in human data. 


\section{Acknowledgment}

This research work was made possible thanks to the financial support from the National Science Foundation (NSF/CMMI-1254906) and the Office of Naval Research (11361538). 

\bibliographystyle{unsrt}
\bibliography{}

\end{document}